\documentclass[letterpaper, 10 pt, conference]{ieeeconf}  % Comment this line out if you need a4paper

\IEEEoverridecommandlockouts                              % This command is only needed if 
\usepackage{graphicx} % for pdf, bitmapped graphics files
\usepackage{amsmath} % assumes amsmath package installed
\usepackage{amssymb}  % assumes amsmath package installed
\usepackage[bookmarks=true]{hyperref}

\usepackage[linesnumbered,ruled,vlined,norelsize]{algorithm2e}

\SetCommentSty{mycommfont}
\usepackage{xcolor}
\definecolor{supergood}{rgb}{.2,.6,.2} 
\definecolor{good}{rgb}{.5,.99,.5} 
\definecolor{equal}{rgb}{0.1,0.1,0.1}  
\definecolor{bad}{rgb}{.9,.7,.1} 
\definecolor{superbad}{rgb}{.9,.1,.1}   
\definecolor{notworking}{rgb}{.5,.5,.5}
\usepackage{array}
\newcolumntype{P}[1]{>{\centering\arraybackslash}p{#1}}

\usepackage{fancyhdr}

\title{\LARGE \bf
Beyond ANN: Exploiting Structural Knowledge\\for Efficient Place Recognition
}

\author{Stefan Schubert, Peer Neubert and Peter Protzel$^{1}$% <-this % stops a space
\thanks{$^{1}$All authors are with Faculty of Electrical Engineering and Automation Technology, Chemnitz University of Technology, Chemnitz, Germany
        {\tt\small \{firstname.lastname\}@etit.tu-chemnitz.de}}%
}

\begin{document}

\maketitle
\thispagestyle{fancy}
\fancyhead[OL]{ 
    \footnotesize
    To appear in Proc. of IEEE International Conference on Robotics and Automation (ICRA), 2021, Xi'an, China. SUBMITTED VERSION\\
    \tiny
    \copyright 2021 IEEE. Personal use of this material is permitted. Permission from IEEE must be obtained for all other uses, in any current or future media, including
    reprinting/republishing this material for advertising or promotional purposes, creating new collective works, for resale or redistribution to servers or lists, or reuse of any copyrighted component of this work in other works.}

%%%%%%%%%%%%%%%%%%%%%%%%%%%%%%%%%%%%%%%%%%%%%%%%%%%%%%%%%%%%%%%%%%%%%%%%%%%%%%%%
\begin{abstract}
Visual place recognition is the task of recognizing same places of query images in a set of database images, despite potential condition changes due to time of day, weather or seasons.
It is important for loop closure detection in SLAM and candidate selection for global localization.
Many approaches in the literature perform computationally inefficient full image comparisons between queries and \textit{all} database images.
There is still a lack of suited methods for efficient place recognition that allow a fast, sparse comparison of only the most promising image pairs without any loss in performance.
While this is partially given by ANN-based methods, they trade speed for precision and additional memory consumption, and many cannot find arbitrary numbers of matching database images in case of loops in the database.
In this paper, we propose a novel fast sequence-based method for efficient place recognition that can be applied online.
It uses relocalization to recover from sequence losses, and exploits usually available but often unused intra-database similarities for a potential detection of \textit{all} matching database images for each query in case of loops or stops in the database.
We performed extensive experimental evaluations over five datasets and 21 sequence combinations, and show that our method outperforms two state-of-the-art approaches and even full image comparisons in many cases, while providing a good tradeoff between performance and percentage of evaluated image pairs.
Source code for Matlab will be provided with publication of this paper.
\end{abstract}

%%%%%%%%%%%%%%%%%%%%%%%%%%%%%%%%%%%%%%%%%%%%%%%%%%%%%%%%%%%%%%%%%%%%%%%%%%%%%%%%
\section{INTRODUCTION}

Visual place recognition is the task of recognizing places of query images $Q$ in a set of database images $DB$, despite potential environmental condition changes due to time of day, weather or seasonal changes.
It is required for loop closure detection in SLAM or candidate selection for global localization.
Visual place recognition can be considered as an embedding of pure image retrieval into mobile robotics.
This allows the exploitation of additional structural knowledge like spatio-temporal sequences in the database and query set.

Many approaches for place recognition perform a full pairwise comparison between all images of database and query, which is not only computationally expensive but potentially error-prone due to the risk of matching similar looking but different places.
To circumvent the computational costs, some approaches use approximate nearest neighbor search (ANN) to find promising candidates for place recognition.
However, ANN-based approaches usually trade computation speed for precision, require additional memory for indexing, and do potentially not find multiple or even all matching database images for a single query image in case of loops or stops in the database.
Finding multiple matches can be important for example in graphSLAM to match a query image not only to the most similar database image but also to another less similar database image with higher pose accuracy.

In this work, we propose a new method for efficient place recognition.
It exploits sequences in database and query to compare only a small fraction of images between database and query, and avoids the usage of ANN-based methods.
In addition, it builds upon yet often unused intra-database similarities $S^{DB}$ for an exploitation of additional inherent structural knowledge without a need for additional sensors.
Intra-database similarities contain information about loops and stops in the database.
In summary, our proposed method 
\begin{itemize}
    \item can be used online
    \item is fast
    \item tremendously reduces the number of image comparisons
    \item preserves or even increases the performance of full image comparisons
    \item handles loops and stops in the database to find multiple matches for each query
    \item can cope with exploration in the query (no matching database images), when other sequence-based methods potentially fail
    \item auto-tunes two thresholds; this allows an application over many datasets with a single set of parameters
\end{itemize}

In the following Sec.~\ref{sec:related_work}, we give an overview over related work.
Next in Sec.~\ref{sec:algo}, we describe our proposed algorithm for efficient place recognition as well as a method for automatic threshold tuning, and describe two relocalization strategies to recover from sequence loss.
Subsequently in Sec.~\ref{sec:results}, we experimentally evaluate our method, compare it to two state of the art methods for sequence-based and ANN-based efficient place recognition, show the benefit of using intra-database similarities, and consider the problem of exploration during the query run and the benefit of relocalization.
Finally, we conclude our work in Sec.~\ref{sec:discussion}.

\section{RELATED WORK}\label{sec:related_work}
Visual place recognition in changing environments is subject of active research.
A more detailed introduction and an overview over existing methods is given in a survey from 2016~\cite{Lowry2016}.
State-of-the-art methods for place recognition build upon deep learning image descriptors.
S\"underhauf et al.~\cite{Suenderhauf2015} showed that the intermediate layers of CNNs trained on an image classification task can serve as a good holistic image descriptor for place recognition.
They identified the conv3-layer of AlexNet~\cite{alexnet} to be robust against condition and viewpoint changes.
NetVLAD~\cite{netvlad} is a holistic image descriptor that was specifically designed and trained for place recognition.
The performance of such holistic descriptors can be further improved by using feature standardization~\cite{Schubert2020}.
Different to holistic image descriptors, CNN-based methods like DELF~\cite{delf} and D2-Net~\cite{d2net} extract local image descriptors for place recognition and localization.
However, some local methods like D2-Net use holistic descriptors for an initial candidate selection due to computational efficiency~\cite{d2net}.

For place recognition, many methods (e.g., \cite{seqSLAM}\cite{Naseer2014}) perform a full pairwise comparison between all database and query descriptors.
However, this approach is computationally inefficient and comes along with a higher risk of comparing similar looking but different places. 
There are at least three classes of methods in place recognition that avoid full image comparisons: sequence-based methods, methods that use approximate nearest neighbor search (ANN), and trained place classifiers.

Most sequence-based methods like SeqSLAM~\cite{seqSLAM}, HMM~\cite{Hansen2014} or MCN~\cite{Neubert2019} are designed to improve the place recognition performance, but are not intended to guide the image comparison for efficiency.
For efficient, sparse image comparisons, Vysotska and Stachniss~\cite{Vysotska2016} proposed the sequence-based method Online Place Recognition (OPR) that performs a graph-based sequence search.
Further, they investigated the incorporation of noisy GPS as location priors to handle loops in the database.
Our approach builds upon sequences for efficiency as well, but performs a more greedy trajectory search for faster computation, while OPR inefficiently expands graph nodes iteratively.
Moreover, OPR suffers under exploratory phases during the query run as illustrated in Fig.~\ref{fig:exploration_problem}.
To circumvent this problem, a relocalization procedure is integrated in our method.
Moreover, we will show that we can handle loops even without GPS by using inherent, often available intra-database similarities $S^{DB}$.

ANN-based approaches for efficient place recognition trade computational speed for precision and additional memory consumption, and cannot find arbitrary numbers of matching database images.
A recent survey~\cite{Li2020} provides a comprehensive comparison of ANN methods on different tasks with an analysis of precision losses.
They identified HNSW~\cite{hnsw} to be particularly suited for high-dimensional data.
FLANN~\cite{flann} obtained low performance for place recognition in~\cite{Vysotska2016}.
MILD~\cite{Han2017} is an ANN-based method specifically designed for place recognition.
ORB-SLAM2~\cite{Mur-Artal2017} uses Bag of Words~\cite{Galvez-Lopez2012} to find nearest neighbors for place recognition.

A third class of methods that avoids full comparisons between database and query images use place classification.
The database is used to train a deep neural network that treats each place as a class, which requires additional and often unavailable knowledge about the database.
\cite{Chen2017} used this approach during training to learn a holistic image descriptor for place recognition.
PlaNet~\cite{Weyand2016} applies this technique for geo-localization also during inference.
For a general overview over global geo-localization techniques please refer to \cite{Brejcha2017}.

Handling loops in the database allows the matching of a query to multiple database images.
For example in graphSLAM~\cite{Grisetti2010}, this allows the connection of a new query node to nodes with lower visual similarity but higher pose accuracy.
Searching multiple matches also increase the risk for wrong matches.
However, since the advent of robust graph optimization for SLAM~\cite{Suenderhauf2012}, wrongly matched places can be robustly detected.
Multiple loop closures are for example used in some graphSLAM approaches (e.g., \cite[p.40]{Grisetti2010}).
MILD~\cite{Han2017} was specifically designed to return multiple loop closure detections for this purpose.

\section{ALGORITHMIC DESCRIPTION}\label{sec:algo}
In this section, we present an efficient place recognition (EPR) for sparse image comparisons between database and query.
EPR is designed to be fast and to work online while it preserves or improves the performance of full image comparisons.
It can handle loops (and stops) in the database to potentially find all matches between database and a query.

The key idea of EPR is the exploitation of spatio-temporal sequences in the database ($DB$) and query ($Q$) set to identify candidates for a subsequent comparison with an incoming query image:
Matching candidates are selected from the $K$ highest similarities of the previous timestep and supplemented by their $v$ successors in $DB$.
These found candidates are used to extract additional similar database candidates from the intra-database similarities $S^{DB}$ to handle loops (and stops) for a final comparison with the new query.

A detailed algorithmic description of EPR is presented in Algorithm~\ref{algo}.
The involved parameters are introduced in Sec.~\ref{sec:params}.
The following Sec.~\ref{sec:overview} provides an overview over the algorithmic steps in Algorithm~\ref{algo}.
Sec.~\ref{sec:tuning} proposes an automatic tuning of the required threshold $\theta_s^{DB}$ and the optional threshold $\theta_s^{reloc}$ in case of event-based relocalization.
Sec.~\ref{sec:reloc} proposes two relocalization strategies to recover from sequence loss.

\subsection{Parameters}\label{sec:params}
EPR requires two parameters and one optional parameter to be set: 
\begin{enumerate}
    \item $K\ldots$ the initial number of matching candidates
    \item $v\ldots$ the number of added subsequent database descriptors of the matching candidates in $DB$. This parameter exploits the sequences in $DB$ and $Q$
    \item (optional) $T^{reloc}\ldots$ timesteps until the next relocalization in case of periodic relocalization (PR). See Sec.~\ref{sec:reloc} for both relocalization strategies.
\end{enumerate}

\subsection{Algorithmic overview}\label{sec:overview}
Input to EPR is an already computed set of image descriptors $db_i \in DB$ and $q_t \in Q$ for every database and new query image.
The algorithm can be roughly divided into the following steps:
\begin{enumerate}
    \item Compute the intra-database similarity $S^{DB}\in\mathbb{R}^{|DB|\times|DB|}$ by comparing all descriptors $db_i\in DB$. For better performance, feature standardization \cite{Schubert2020} over all $db_i$ could be applied (line 1) 
    \item Perform a relocalization for the first incoming query descriptor $q_1$ (line 2)
    \item According to Sec.~\ref{sec:tuning}, auto-tune $\theta_s^{DB}$ (and $\theta_s^{reloc}$ in case of event-based relocalization (ER), see Sec.~\ref{sec:reloc}) (lines 3-4)
    \item For every new incoming query descriptor $q_t$ at timestep $t$ (line 5):
    \begin{enumerate}
        \item Select matching candidates $db_c$ with $c\in C^{DB}$ based on the $K$-best comparisons from the previous timestep $t-1$, their $v$ successors in $DB$ and their most similar database images with $S^{DB}\geq\theta_s^{DB}$ (lines 6-10)
        \item Compare all matching candidates $db_c$ with $c\in C^{DB}$ to the current query descriptor $q_t$ (line 11)
        \item If $relocalize(\cdot) \leftrightarrow true$ (see Sec.~\ref{sec:reloc}), perform a relocalization by comparing query $q_t$ to all database descriptors $db_i\in DB$ (lines 12-13)
        \item Else if $relocalize(\cdot) \leftrightarrow false$, find the database descriptors $c$ with the $K$-highest similarities $S_{ct}$ for the current timestep $t$ and add their most similar database descriptors $\hat{c}$ if $S^{DB}_{\hat{c}c} \geq \theta_s^{DB}$. Compare all found candidates to the current query $q_t$ to compute similarities $S_{ct}$ (lines 14-18)
    \end{enumerate}
\end{enumerate}

\SetInd{0.4em}{0.6em}
\begin{algorithm}[t]
    \caption{Algorithmic description of our proposed Efficient Place Recognition (EPR) method}
    \label{algo}
    
    \small
    
    \KwData{$db_i\in DB$, $q_t\in Q$}
    \KwIn{parameters $K$, $v$, ($T^{reloc}$ in case of periodic relocalization, see Sec.~\ref{sec:reloc}}
    \KwResult{$S\in\mathbb{R}^{|DB|\times|Q|}$}
    
    \smallskip\tcp{compute intra-database similarities $S^{DB}$ from (standardized) database descriptors}
    $S^{DB}_{ij} = \frac{db^T_i \cdot db_j}{\|db_i\|\cdot \|db_j\|} \ \ \ \forall i,j = 1,\ldots,|DB|$
    
    \smallskip\tcp{initial relocalization}
    $S_{i1} = \frac{db_i^T \cdot q_1}{\|db_i\|\cdot\|q_1\|}  \ \ \ \forall i=1,\ldots,|DB|$

    \smallskip\tcp{auto-tune $\theta_s^{DB}$ (and $\theta_s^{reloc}$) according to Sec.~\ref{sec:tuning}}
    $\theta_s^{DB} = autotune(S^{DB})$
    
    $\theta_s^{reloc} = autotune(S_{:1})$

    \smallskip\tcp{for all incoming query images}
    \For{$t=2,\ldots,|Q|$}{
        \smallskip\tcp{find $K$ highest similarities from previous timestep $t{-}1$}
        $C^{DB} := \underset{c}{\text{K-argmax}}(S_{c,(t-1)}, \ K)$
        
        \smallskip\tcp{for all candidates in $C^{DB}$ find database images in the intra-database similarities $S^{DB}$ with similarity $\geq \theta_s^{DB}$}
        \For{$\forall c\in C^{DB}$}{
            $C^{DB} := C^{DB} \cup \{\hat{c} \mid S^{DB}_{\hat{c}c} \geq \theta_s^{DB}, \hat{c}\notin C^{DB}\}$
        }
    
        \smallskip\tcp{exploit the sequence assumption by adding the $v$ consecutive images in $DB$ for all candidates in $C^{DB}$}
        \For{$\forall c\in C^{DB}$}{
            $C^{DB} := C^{DB} \cup \{c+i, \ \forall i=1,\ldots,v \mid c+i\notin C^{DB}\}$
        }
        
        \smallskip\tcp{compare query $q_t$ to all candidates in $C^{DB}$}
        $S_{ct} = \frac{db_c^T \cdot q_t}{\|db_c\|\cdot\|q_t\|} \ \ \ \forall c\in C^{DB}$
        
        \smallskip\tcp{periodic or event-based relocalizaiton given either $\theta_s^{reloc}$ or $T^{reloc}$ (see Sec.~\ref{sec:reloc})}
        \eIf{relocalize($\cdot$)}{
            $S_{it} = \frac{db_i^T \cdot q_t}{\|db_i\|\cdot\|q_t\|} \ \ \ \forall i=1,\ldots,|DB|$
        }{
            \smallskip\tcp{find loops in intra-database similarities $S^{DB}$ for $K$ highest similarities in current timestep}
            $C^{DB} := \underset{c}{\text{K-argmax}}(S_{ct}, \ K)$
            
            \For{$\forall c\in C^{DB}$}{
                $C^{DB} := C^{DB} \cup \{\hat{c} \mid S^{DB}_{\hat{c}c} \geq \theta_s^{DB}, \hat{c}\notin C^{DB}\}$
            }
            
            $S_{ct} = \frac{db_c^T \cdot q_t}{\|db_c\|\cdot\|q_t\|} \ \ \ \forall c\in C^{DB}$ 
        }
    }
\end{algorithm}

\subsection{Automatic tuning of $\theta_s^{DB}$ and $\theta_s^{reloc}$}\label{sec:tuning}
The two thresholds $\theta_s^{DB}$ and $\theta_s^{reloc}$ in Algorithm~\ref{algo} are particularly dependent on the dataset and the used image descriptor.
In order to apply EPR over a wider range of datasets with a single set of parameters, both thresholds have to be tuned automatically.

\begin{figure}[t]
    \centering
    \includegraphics[width=0.49\linewidth]{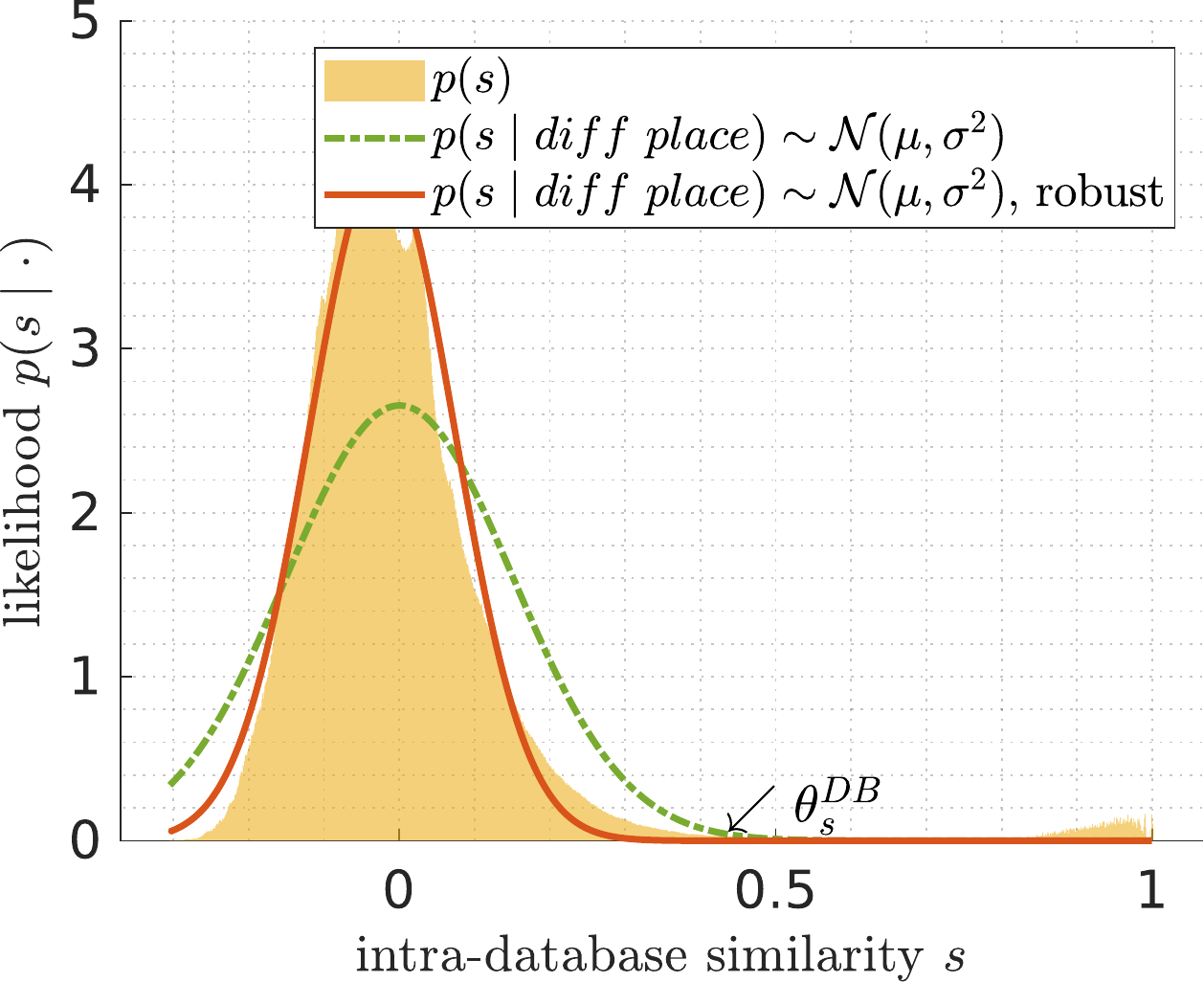}
    \includegraphics[width=0.49\linewidth]{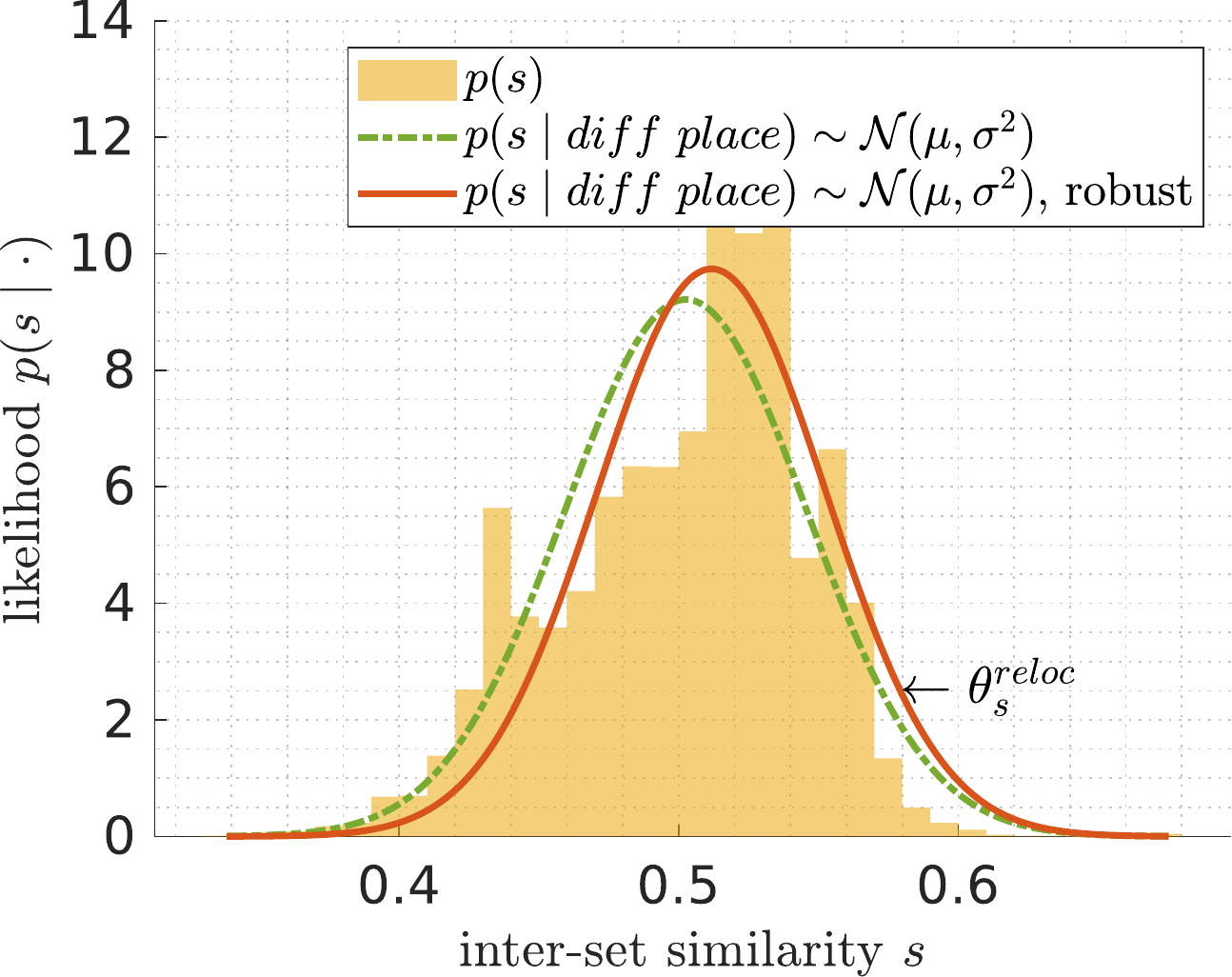}

    \vspace{-0.2cm}
    
        \caption{Distribution of intra-database similarities $S^{DB}$ (left) and of similarities $S_{:1}$ between all database images and the first query image (right). The solid red lines show the robustly fitted normal distribution model and the dashed green lines the non-robustly fitted normal distribution model. The parameters $\theta_s^{DB}$ and $\theta_s^{reloc}$ were finally determined from the red line. See Sec.~\ref{sec:tuning} for a detailed description.} %Nordland spring-winter
    \label{fig:tuning}
    
    \vspace{-0.2cm}
\end{figure}

Our proposed unsupervised method tries to model the distribution of similarities of different places to exclude them from all similarities with a certain probability.
As we cannot divide similarities from descriptor comparisons into same and different places without labels, we assume that there are more different places than same places.
This assumption is usually valid for larger-scale datasets.
Fig.~\ref{fig:tuning} shows a typical distribution of the intra-database similarities $S^{DB}$ and the similarities $S_{:1}$ between database and the first query.

The automatic parameter tuning fits a normal distribution 
\begin{align}
    \mathcal{N}(\mu, \sigma^2)\sim p(s\mid \text{different places}) \label{eq:model}
\end{align}
to the similarity values.
Because of the contained outliers (the same places) in $S^{DB}$ and $S_{:1}$, a robust parameter estimation~\cite{MAD} for normal distributions is used for $S^{DB}$
\begin{align}
    \mu &\approx median(S^{DB}) \label{eq:mu}\\ 
    \sigma &\approx median(|S^{DB} - median(S^{DB})|) / 0.675 \label{eq:sig}
\end{align}
and similarly for $S_{:1}$.
For $\sigma$ we use the normalized median absolute deviation (MADN)~\cite{MAD}.

Given the model in Eq.~\ref{eq:model} and the estimated parameters from Eq.~\ref{eq:mu} and \ref{eq:sig}, we determine the parameters $\theta_s^{DB}$ and $\theta_s^{reloc}$ that satisfy
\begin{align}
    p(s< \theta_s^{DB}\mid \text{different places}) &= 1 - 10^{-6} \\
    p(s< \theta_s^{reloc}\mid \text{different places}) &= 0.95
\end{align}
Similarities above these thresholds are likely to belong to same places.
We can use a higher probability for the intra-database similarities, since the place descriptors perform better under more constant conditions.

\begin{figure}[t]
    \centering
    \includegraphics[width=0.25\linewidth]{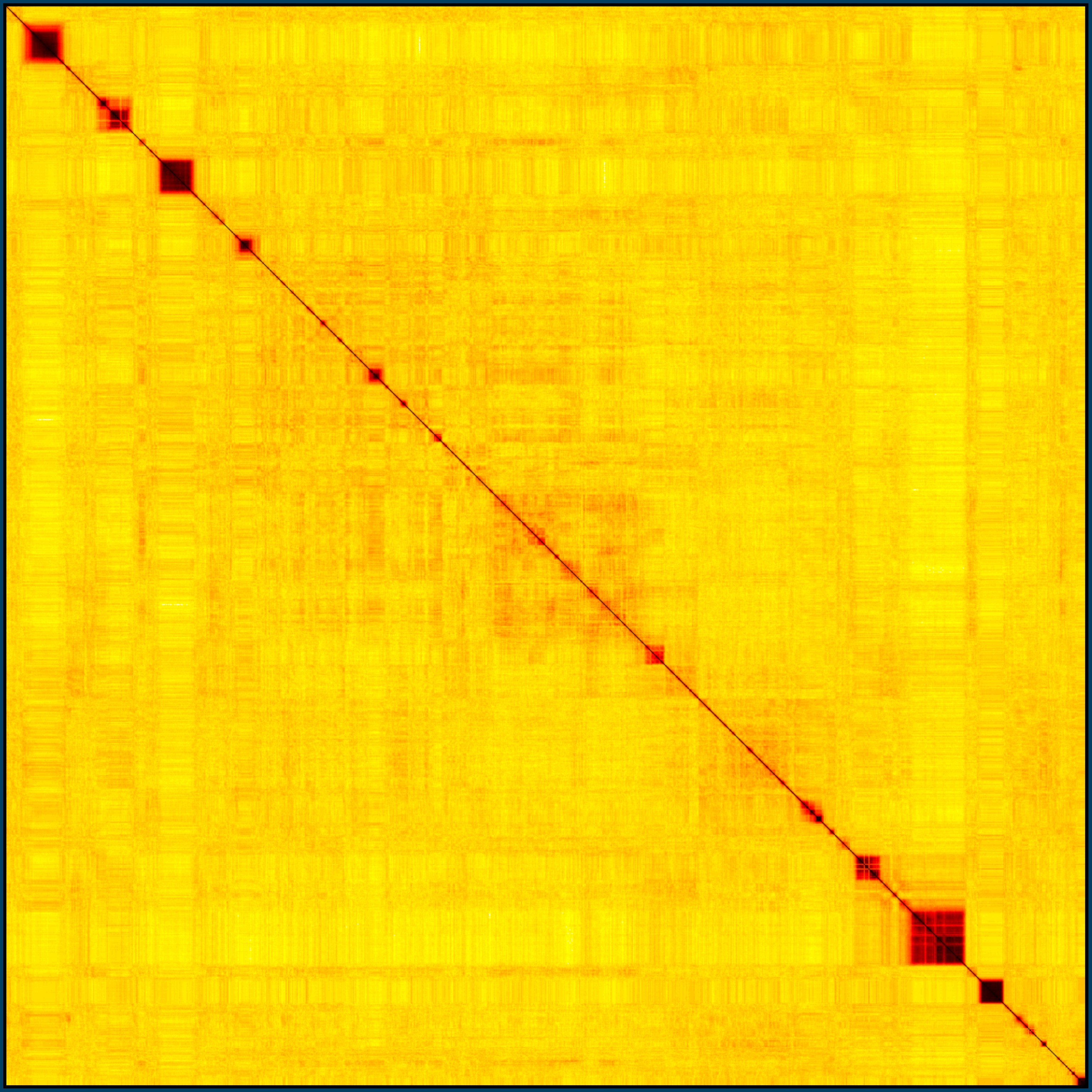}
    \includegraphics[width=0.25\linewidth]{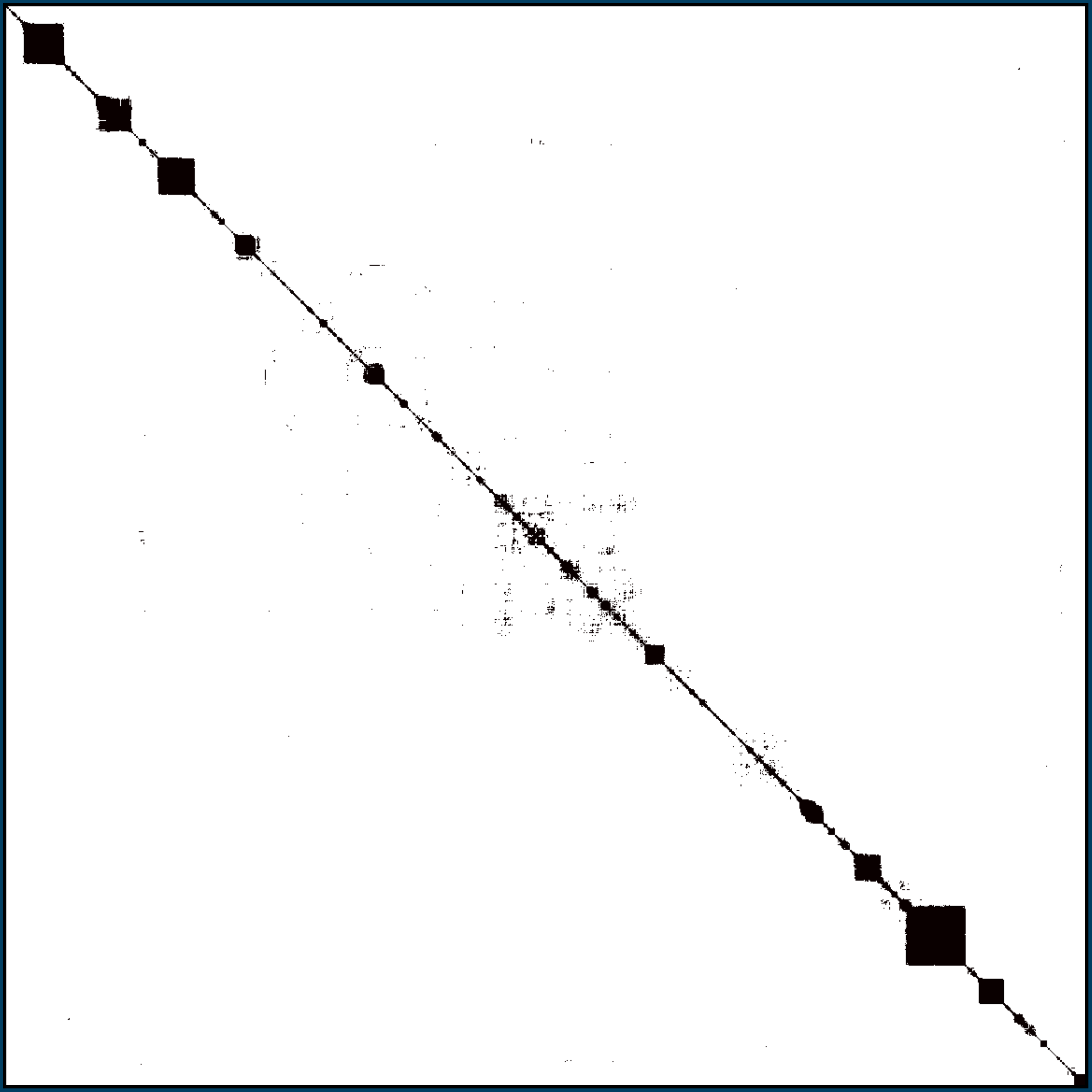}
    
    \vspace{-0.25cm}
    
    \caption{Intra-database similarities $S^{DB}$ (left) and thresholded with the auto-tuned $\theta_s^{DB}$ according to Sec.~\ref{sec:tuning} (right)} %from CMU 20110421-20100901
    \label{fig:SDB}
    
    \vspace{-0.35cm}
\end{figure}

Fig.~\ref{fig:tuning} compares the result of a robust and regular normal distribution fitting.
While the robust fitting performs well, the regular fitting (with mean and standard deviation) rather fails.
Fig.~\ref{fig:SDB} shows an intra-database similarity matrix $S^{DB}$ thresholded with the auto-tuned $\theta_s^{DB}$.

\subsection{Relocalization strategy}\label{sec:reloc}
EPR performs image comparisons locally in the similarity matrix $S$: Candidate descriptors in the database for the current timestep $t$ are selected from neighboring descriptors (with similar indices) from the previous timestep.
Merely the intra-database similarities allow the choice of more distant candidates (much higher or smaller indices) in the database.
While this sequence-based approach allows a sparse image comparison, it can also cause sequence loss, especially after exploration during the query run.
In case of sequence loss, new query images could simply be compared to non-matching database images despite existing matching database images.
This behavior can be observed in Fig.~\ref{fig:exploration_problem}.

In order to recover from a sequence loss, we use a relocalization that compares the current query $q_t$ to all database descriptors $db_i\in DB$ with
\begin{align}
    S_{it} = \frac{db_i^T \cdot q_t}{\|db_i\|\cdot\|q_t\|} \ \ \ \forall i=1,\ldots,|DB|
\end{align}
To be more efficient than this full comparison, ANN-based methods could be used.
However, these methods would involve additional properties that have to be evaluated, and are therefore subject of future work.

We propose two relocalization strategies: 1) periodic relocalization and 2) event-based relocalization.
The qualitative performance of both strategies on a single dataset can be seen in Fig.~\ref{fig:exploration_problem}.

\subsubsection{Periodic relocalization (PR)}
A relocalization is performed periodically every $T^{reloc}$ timesteps with
\begin{align}
    relocalize(\cdot) = \begin{cases}
        true\text{, if } t\bmod{}T^{reloc} \leftrightarrow 0
        \\
        false\text{, otherwise}
    \end{cases}
\end{align}

\subsubsection{Event-based relocalization (ER)}
Given the already computed similarities $S_{it}$ for the current timestep $t$, a relocalization is performed if none of the similarities exceed threshold $\theta_s^{reloc}$ with
\begin{align}
    relocalize(\cdot) = \begin{cases}
        true\text{, if } \nexists i \text{ s.t. } S_{it}\geq \theta_s^{reloc}
        \\
        false\text{, otherwise}
    \end{cases}
\end{align}

\section{EXPERIMENTAL RESULTS}\label{sec:results}

\subsection{Experimental setup}
NetVLAD~\cite{netvlad} and AlexNet~\cite{alexnet} are used as CNN-descriptors in the following experiments.
For NetVLAD, we use the author's implementation trained on the Pitts30k dataset with VGG-16 and whitening that returns a normalized $4096$-dimensional descriptor.
For AlexNet, we use the flattened $64896$-dimensional conv3-layer output of Matlab's ImageNet model.
The area under the precision-recall curve (AUC) is used as performance metric over all experiments.
The cosine similarity is used to measure the similarity between two descriptors.

For EPR, we used the parameters $K=5$, $v=5$ and $T^{reloc}=100$ for PR.
Feature standardization~\cite{Schubert2020} on the database descriptors was used before the computation of intra-database similarities.

EPR is compared to OPR and HNSW:
1) Online Place Recognition (OPR)~\cite{Vysotska2016} is a sequence-based approach.
We used the author's C++ implementation with parameters $K=5$ and $\alpha=0.6$.
2) Hierarchical Navigable Small World graphs (HNSW)~\cite{hnsw} is an ANN method that was shown to be particularly suited for high-dimensional data~\cite{Li2020}.
We used the author's C++ implementation with parameters $ef=20$ and $M=40$, and search for $K=5$ nearest neighbors.
AlexNet descriptors were downsampled to $6490$ dimensions before HNSW using Gaussian random projection.

\subsection{Datasets}
Our evaluation is based on five datasets with different characteristics regarding environment, appearance changes, single or multiple visits of places, possible stops, or viewpoint changes.
We basically used the same five datasets \textbf{Nordland}~\cite{ds_nordland}, \textbf{StLucia} (Various Times of the Day)~\cite{ds_stlucia}, \textbf{CMU} Visual Localization~\cite{ds_cmu}, \textbf{Gardens Point} Walking~\cite{ds_gardenspoint_riverside}, and \textbf{Oxford} RobotCar~\cite{ds_robotcar} as described in our previous publication~\cite{Neubert2019}.

For Oxford, we use a new set of sequences sampled at 1Hz with the recently published accurate ground truth data~\cite{rtk}.
For Nordland, we sampled the first two hours of each season including stops at 1Hz and removed tunnels.

\subsection{Qualitative performance}
\begin{figure}[t]
    \centering
    \includegraphics[width=0.325\linewidth]{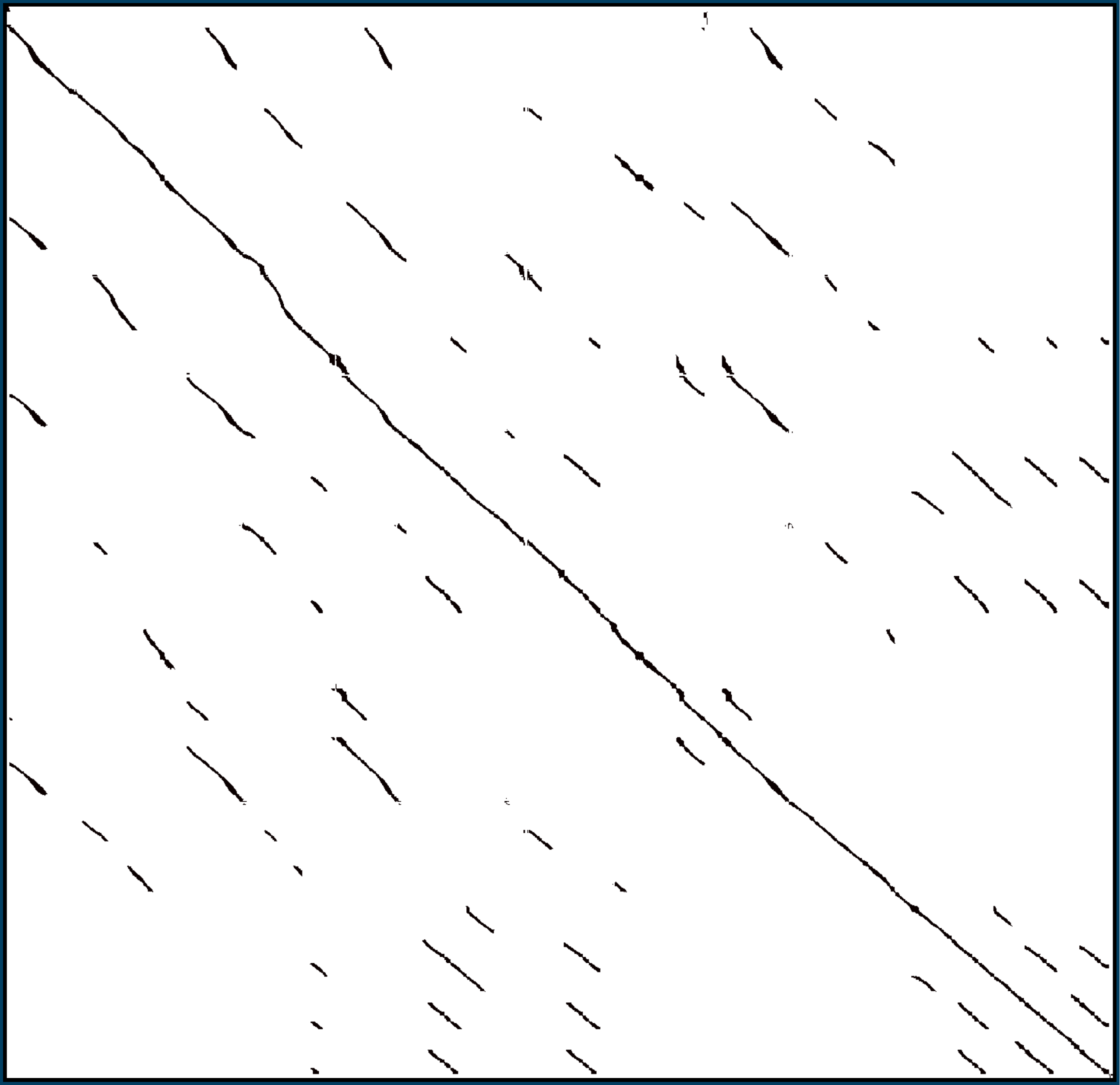}
    \includegraphics[width=0.325\linewidth]{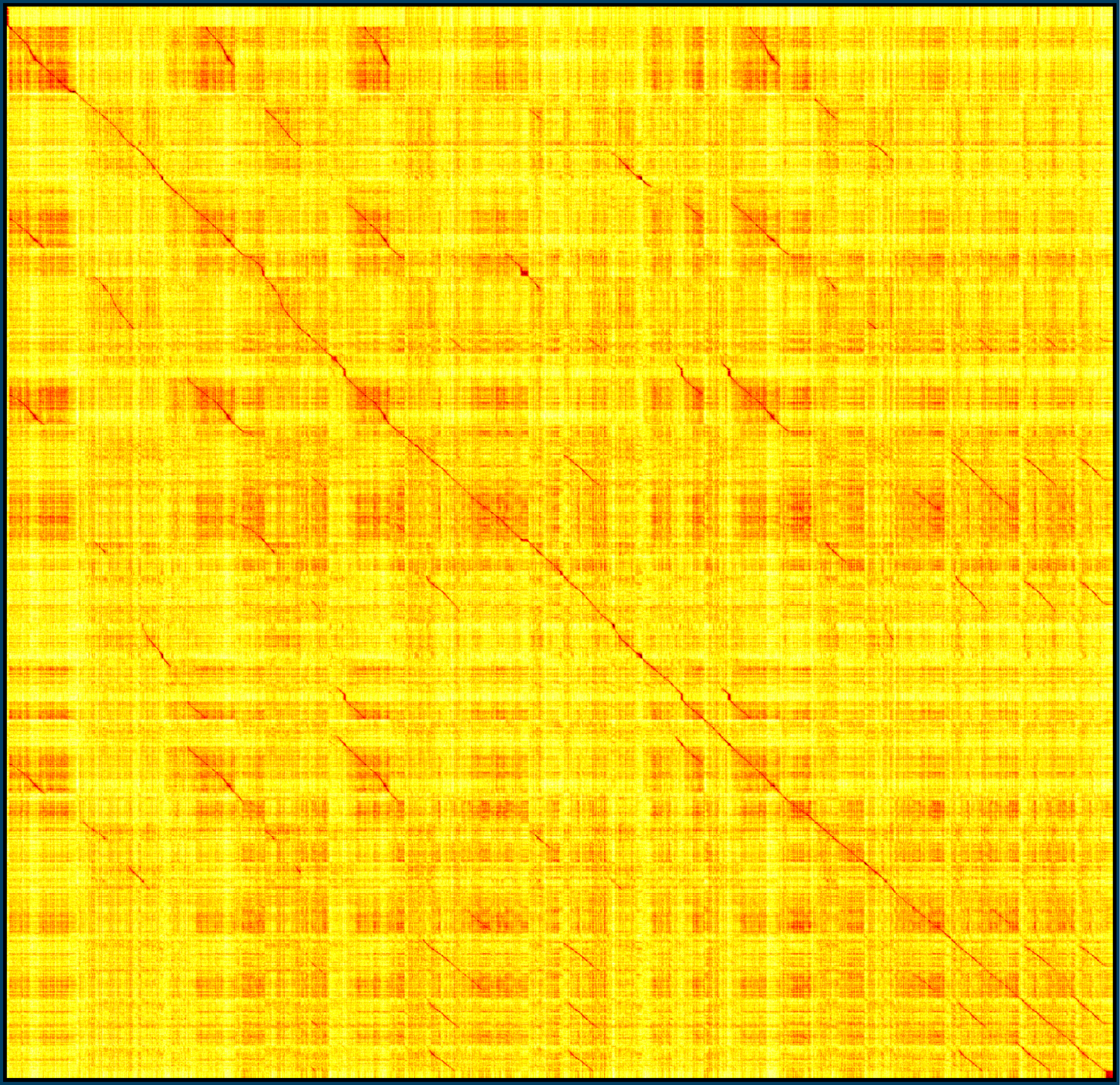}
    \includegraphics[width=0.325\linewidth]{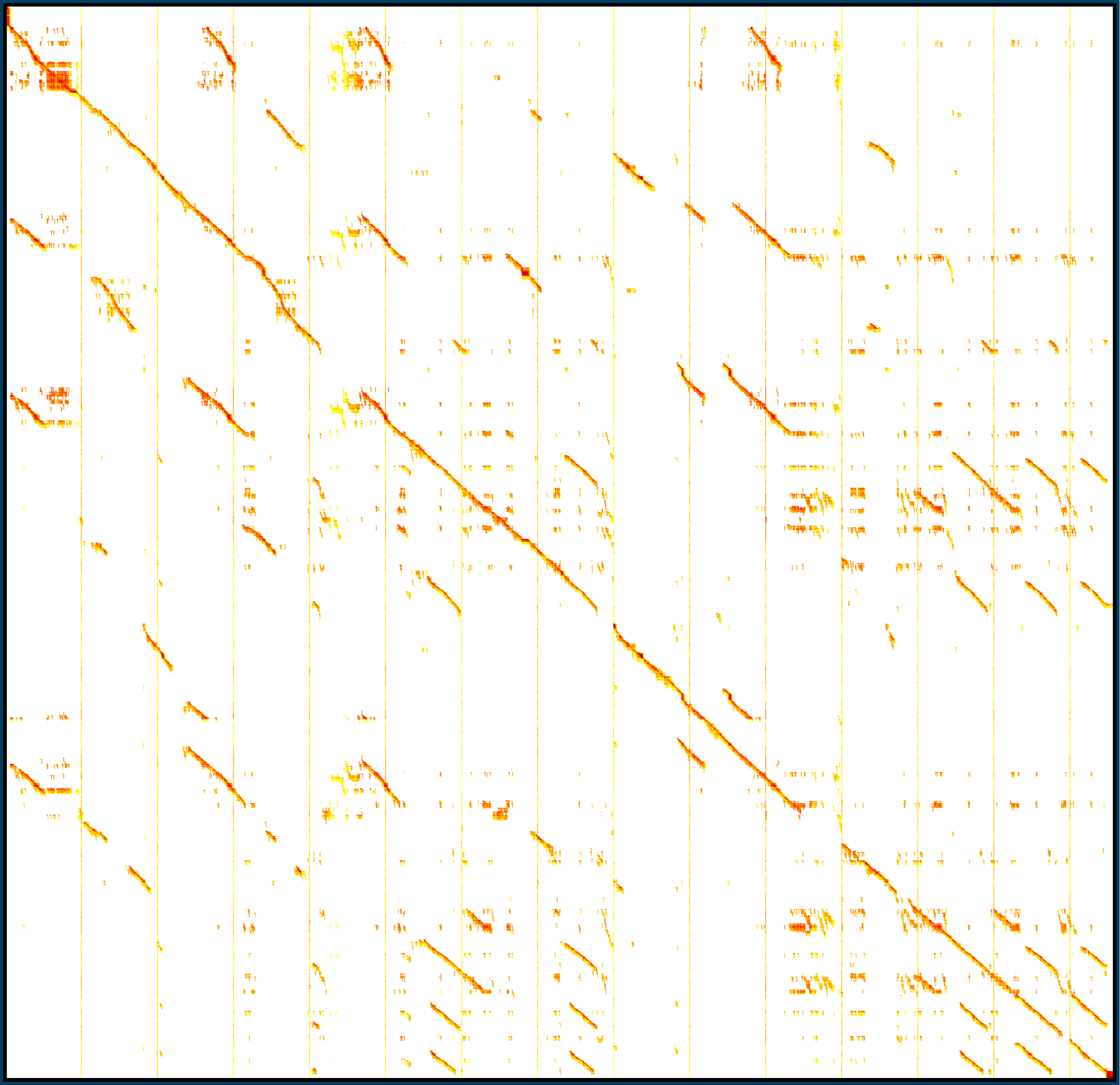}

\vspace{-0.1cm}

    \caption{Evaluated similarities $S\in\mathbb{R}^{|DB|\times|Q|}$ between database and query images for StLucia 08:45. White pixels represent uncompared image pairs. (left) ground truth, (mid) full image comparisons, (right) EPR-PR}
    \label{fig:results_overall}
    
    \vspace{-0.2cm}
\end{figure}

Fig.~\ref{fig:results_overall} visualizes the qualitative performance of EPR with periodic relocalization (PR) together with ground truth and full image comparisons.
The sparse similarity matrix $S$ from EPR-PR clearly emphasizes the high efficiency compared to full image comparisons and its ability to correctly match multiple database images to a current query.
The periodic relocalization can be seen as thin vertical lines.

\subsection{Performance comparison}
To measure the quantitative performance of our method EPR with periodic (PR) and event-based (ER) relocalization, we conduced experiments over five datasets with 21 sequence combinations.
We compare it with the \textit{full} image comparisons and with OPR and HNSW.
We measured the single-matching performance which evaluates for each query image only the highest similarity to a database image as well as the multi-matching performance that evaluates all similarities from all compared image pairs per query image.
The multi-matching performance evaluates the capability of a method to find multiple matches for a query in case of loops in the database.

Table~\ref{tab:AUC} shows the achieved absolute performance for each evaluated method.
Fig.~\ref{fig:density-vs-AUC} illustrates their relation to the percentage of evaluated image pairs.
The single-matching performance of the full image comparisons can be maintained by most methods.
Only OPR performs worse on approximately the half of the datasets, because it often loses track of the sequences due to low similarities, exploration during the query run and stops in the database (see also Sec.~\ref{sec:reloc_problem}).
EPR-ER with AlexNet fails on a single Oxford dataset with query images at night, since it cannot infer the need for a relocalization.
\boldmath
\begin{table*}[t]
\centering
\caption{Performance (AUC) of the evaluated algorithms with NetVLAD or AlexNet descriptor. Best values per row and descriptor are bold. The colored arrows indicate large ($\ge$25\% better/worse) or medium ($\ge$10\%) deviation compared to ``full''.}\vspace{-0.3cm}

\label{tab:AUC}
\resizebox{1\textwidth}{!}{%
\begin{tabular}{lll||c|cc|P{1.25cm}|cc||c|cc|P{1.25cm}|cc}
	\hline
	                 &                     &                     &                                                                                                               \multicolumn{6}{c||}{\textbf{NetVLAD}}                                                                                                                &                                                                                                                \multicolumn{6}{c}{\textbf{AlexNet}}                                                                                                                \\
	\textbf{Dataset} & \textbf{Database}   & \textbf{Query}      &         \textbf{full}         &               \textbf{EPR-PR}                               &               \textbf{EPR-ER} & \textbf{EPR-PR w/o $S^{DB}$}              &                \textbf{OPR}                 &                \textbf{HNSW}                &         \textbf{full}         &               \textbf{EPR-PR}                               &               \textbf{EPR-ER}  & \textbf{EPR-PR w/o $S^{DB}$}             &                \textbf{OPR}                &                \textbf{HNSW}                \\ \hline\hline
	                 &                     &                     &                                                                                                                                                                                                                                        \multicolumn{12}{c}{\textbf{Single-Matching Performance}}                                                                                                                                                                                                                                         \\ \hline
	Nordland         & fall                & spring              &     0.74 \color{equal}{}      &      0.72 \color{equal}{$\rightarrow$}      &      0.73 \color{equal}{$\rightarrow$}      & \textbf{0.90} \color{good}{$\nearrow$}      &        0.61 \color{bad}{$\searrow$}         &      0.69 \color{equal}{$\rightarrow$}      & \textbf{1.00} \color{equal}{} & \textbf{1.00} \color{equal}{$\rightarrow$}  & \textbf{1.00} \color{equal}{$\rightarrow$}  & \textbf{1.00} \color{equal}{$\rightarrow$}  &     0.93 \color{equal}{$\rightarrow$}      &      0.99 \color{equal}{$\rightarrow$}      \\
	                 & fall                & winter              &     0.68 \color{equal}{}      &      0.68 \color{equal}{$\rightarrow$}      &      0.67 \color{equal}{$\rightarrow$}      & \textbf{0.87} \color{supergood}{$\uparrow$} &     0.50 \color{superbad}{$\downarrow$}     &      0.65 \color{equal}{$\rightarrow$}      &     0.95 \color{equal}{}      &      0.96 \color{equal}{$\rightarrow$}      & \textbf{0.97} \color{equal}{$\rightarrow$}  & \textbf{0.97} \color{equal}{$\rightarrow$}  &     0.91 \color{equal}{$\rightarrow$}      &      0.93 \color{equal}{$\rightarrow$}      \\
	                 & spring              & winter              &     0.75 \color{equal}{}      &      0.75 \color{equal}{$\rightarrow$}      &      0.71 \color{equal}{$\rightarrow$}      & \textbf{0.85} \color{good}{$\nearrow$}      &        0.57 \color{bad}{$\searrow$}         &      0.72 \color{equal}{$\rightarrow$}      &     0.95 \color{equal}{}      &      0.94 \color{equal}{$\rightarrow$}      &      0.98 \color{equal}{$\rightarrow$}      & 0.96 \color{equal}{$\rightarrow$}           & \textbf{1.00} \color{equal}{$\rightarrow$} &      0.94 \color{equal}{$\rightarrow$}      \\
	                 & summer              & spring              &     0.76 \color{equal}{}      &      0.75 \color{equal}{$\rightarrow$}      &      0.76 \color{equal}{$\rightarrow$}      & \textbf{0.90} \color{good}{$\nearrow$}      &     0.50 \color{superbad}{$\downarrow$}     &      0.72 \color{equal}{$\rightarrow$}      &     0.99 \color{equal}{}      & \textbf{1.00} \color{equal}{$\rightarrow$}  & \textbf{1.00} \color{equal}{$\rightarrow$}  & \textbf{1.00} \color{equal}{$\rightarrow$}  & \textbf{1.00} \color{equal}{$\rightarrow$} & 0.99 \color{equal}{$\rightarrow$}  \\
	                 & summer              & fall                &     0.96 \color{equal}{}      &      0.96 \color{equal}{$\rightarrow$}      &      0.96 \color{equal}{$\rightarrow$}      & \textbf{1.00} \color{equal}{$\rightarrow$}  &      0.91 \color{equal}{$\rightarrow$}      &      0.95 \color{equal}{$\rightarrow$}      & \textbf{1.00} \color{equal}{} & \textbf{1.00} \color{equal}{$\rightarrow$}  & \textbf{1.00} \color{equal}{$\rightarrow$}  & \textbf{1.00} \color{equal}{$\rightarrow$}  & \textbf{1.00} \color{equal}{$\rightarrow$} & \textbf{1.00} \color{equal}{$\rightarrow$}  \\
	StLucia          & 100909-0845         & 190809-0845         &     0.95 \color{equal}{}      & \textbf{0.96} \color{equal}{$\rightarrow$}  & \textbf{0.96} \color{equal}{$\rightarrow$}  & 0.95 \color{equal}{$\rightarrow$}           & \textbf{0.96} \color{equal}{$\rightarrow$}  & 0.95 \color{equal}{$\rightarrow$}  & \textbf{0.99} \color{equal}{} & \textbf{0.99} \color{equal}{$\rightarrow$}  & \textbf{0.99} \color{equal}{$\rightarrow$}  & 0.98 \color{equal}{$\rightarrow$}           & \textbf{0.99} \color{equal}{$\rightarrow$} & \textbf{0.99} \color{equal}{$\rightarrow$}  \\
	                 & 100909-1000         & 210809-1000         &     0.96 \color{equal}{}      &      0.96 \color{equal}{$\rightarrow$}      & \textbf{0.97} \color{equal}{$\rightarrow$}  & 0.93 \color{equal}{$\rightarrow$}           &      0.95 \color{equal}{$\rightarrow$}      &      0.95 \color{equal}{$\rightarrow$}      & \textbf{0.98} \color{equal}{} & \textbf{0.98} \color{equal}{$\rightarrow$}  & \textbf{0.98} \color{equal}{$\rightarrow$}  & \textbf{0.98} \color{equal}{$\rightarrow$}  & \textbf{0.98} \color{equal}{$\rightarrow$} & \textbf{0.98} \color{equal}{$\rightarrow$}  \\
	                 & 100909-1210         & 210809-1210         &     0.97 \color{equal}{}      & \textbf{0.98} \color{equal}{$\rightarrow$}  & \textbf{0.98} \color{equal}{$\rightarrow$}  & \textbf{0.98} \color{equal}{$\rightarrow$}  &      0.89 \color{equal}{$\rightarrow$}      & 0.97 \color{equal}{$\rightarrow$}  & \textbf{1.00} \color{equal}{} & \textbf{1.00} \color{equal}{$\rightarrow$}  & \textbf{1.00} \color{equal}{$\rightarrow$}  & 0.99 \color{equal}{$\rightarrow$}           &     0.97 \color{equal}{$\rightarrow$}      & \textbf{1.00} \color{equal}{$\rightarrow$}  \\
	                 & 100909-1410         & 190809-1410         &     0.95 \color{equal}{}      & \textbf{0.97} \color{equal}{$\rightarrow$}  & \textbf{0.97} \color{equal}{$\rightarrow$}  & 0.91 \color{equal}{$\rightarrow$}           & \textbf{0.97} \color{equal}{$\rightarrow$}  & 0.95 \color{equal}{$\rightarrow$}  & \textbf{0.99} \color{equal}{} & \textbf{0.99} \color{equal}{$\rightarrow$}  & \textbf{0.99} \color{equal}{$\rightarrow$}  & \textbf{0.99} \color{equal}{$\rightarrow$}  & \textbf{0.99} \color{equal}{$\rightarrow$} & \textbf{0.99} \color{equal}{$\rightarrow$}  \\
	                 & 110909-1545         & 180809-1545         &     0.92 \color{equal}{}      & \textbf{0.94} \color{equal}{$\rightarrow$}  & \textbf{0.94} \color{equal}{$\rightarrow$}  & 0.91 \color{equal}{$\rightarrow$}           &        0.82 \color{bad}{$\searrow$}         & 0.92 \color{equal}{$\rightarrow$}  & \textbf{0.99} \color{equal}{} & \textbf{0.99} \color{equal}{$\rightarrow$}  & \textbf{0.99} \color{equal}{$\rightarrow$}  & 0.98 \color{equal}{$\rightarrow$}           &     0.98 \color{equal}{$\rightarrow$}      & \textbf{0.99} \color{equal}{$\rightarrow$}  \\
	CMU              & 20110421            & 20100901            & \textbf{0.97} \color{equal}{} & \textbf{0.97} \color{equal}{$\rightarrow$}  &      0.96 \color{equal}{$\rightarrow$}      & \textbf{0.97} \color{equal}{$\rightarrow$}  &        0.85 \color{bad}{$\searrow$}         & \textbf{0.97} \color{equal}{$\rightarrow$}  &     0.95 \color{equal}{}      &      0.96 \color{equal}{$\rightarrow$}      &      0.94 \color{equal}{$\rightarrow$}      & \textbf{0.97} \color{equal}{$\rightarrow$}  &    0.62 \color{superbad}{$\downarrow$}     & 0.95 \color{equal}{$\rightarrow$}  \\
	                 & 20110421            & 20100915            & \textbf{0.97} \color{equal}{} &      0.96 \color{equal}{$\rightarrow$}      & \textbf{0.97} \color{equal}{$\rightarrow$}  & 0.96 \color{equal}{$\rightarrow$}           &        0.81 \color{bad}{$\searrow$}         & \textbf{0.97} \color{equal}{$\rightarrow$}  &     0.96 \color{equal}{}      & \textbf{0.97} \color{equal}{$\rightarrow$}  & \textbf{0.97} \color{equal}{$\rightarrow$}  & \textbf{0.97} \color{equal}{$\rightarrow$}  &        0.78 \color{bad}{$\searrow$}        & 0.96 \color{equal}{$\rightarrow$}  \\
	                 & 20110421            & 20101221            &     0.92 \color{equal}{}      & \textbf{0.98} \color{equal}{$\rightarrow$}  &      0.92 \color{equal}{$\rightarrow$}      & \textbf{0.98} \color{equal}{$\rightarrow$}  &      0.96 \color{equal}{$\rightarrow$}      &      0.91 \color{equal}{$\rightarrow$}      &     0.93 \color{equal}{}      &      0.97 \color{equal}{$\rightarrow$}      &      0.97 \color{equal}{$\rightarrow$}      & \textbf{0.98} \color{equal}{$\rightarrow$}  &     0.92 \color{equal}{$\rightarrow$}      & 0.93 \color{equal}{$\rightarrow$}  \\
	                 & 20110421            & 20110202            & \textbf{0.99} \color{equal}{} & \textbf{0.99} \color{equal}{$\rightarrow$}  & \textbf{0.99} \color{equal}{$\rightarrow$}  & \textbf{0.99} \color{equal}{$\rightarrow$}  &        0.83 \color{bad}{$\searrow$}         & \textbf{0.99} \color{equal}{$\rightarrow$}  & \textbf{0.98} \color{equal}{} &      0.97 \color{equal}{$\rightarrow$}      & \textbf{0.98} \color{equal}{$\rightarrow$}  & 0.93 \color{equal}{$\rightarrow$}           &     0.90 \color{equal}{$\rightarrow$}      & \textbf{0.98} \color{equal}{$\rightarrow$}  \\
	GardensPoint     & day-right           & day-left            & \textbf{1.00} \color{equal}{} & \textbf{1.00} \color{equal}{$\rightarrow$}  & \textbf{1.00} \color{equal}{$\rightarrow$}  & \textbf{1.00} \color{equal}{$\rightarrow$}  & \textbf{1.00} \color{equal}{$\rightarrow$}  & \textbf{1.00} \color{equal}{$\rightarrow$}  &     0.98 \color{equal}{}      & \textbf{1.00} \color{equal}{$\rightarrow$}  & \textbf{1.00} \color{equal}{$\rightarrow$}  & \textbf{1.00} \color{equal}{$\rightarrow$}  & \textbf{1.00} \color{equal}{$\rightarrow$} & 0.98 \color{equal}{$\rightarrow$}  \\
	                 & day-right           & night-right         &     0.97 \color{equal}{}      &      0.99 \color{equal}{$\rightarrow$}      &      0.99 \color{equal}{$\rightarrow$}      & \textbf{1.00} \color{equal}{$\rightarrow$}  & \textbf{1.00} \color{equal}{$\rightarrow$}  & 0.97 \color{equal}{$\rightarrow$}  &     0.96 \color{equal}{}      &      0.98 \color{equal}{$\rightarrow$}      &      0.97 \color{equal}{$\rightarrow$}      & \textbf{1.00} \color{equal}{$\rightarrow$}  &     0.97 \color{equal}{$\rightarrow$}      & 0.96 \color{equal}{$\rightarrow$}  \\
	                 & day-left            & night-right         &     0.94 \color{equal}{}      &      0.97 \color{equal}{$\rightarrow$}      &      0.96 \color{equal}{$\rightarrow$}      & \textbf{1.00} \color{equal}{$\rightarrow$}  &      0.98 \color{equal}{$\rightarrow$}      &      0.93 \color{equal}{$\rightarrow$}      &     0.76 \color{equal}{}      &        0.92 \color{good}{$\nearrow$}        &   \textbf{0.93} \color{good}{$\nearrow$}    & 0.92 \color{good}{$\nearrow$}               &   \textbf{0.93} \color{good}{$\nearrow$}   & 0.77 \color{equal}{$\rightarrow$}  \\
	Oxford           & 2014-12-09-13-21-02 & 2015-05-19-14-06-38 & \textbf{1.00} \color{equal}{} & \textbf{1.00} \color{equal}{$\rightarrow$}  & \textbf{1.00} \color{equal}{$\rightarrow$}  & 0.99 \color{equal}{$\rightarrow$}           &      0.99 \color{equal}{$\rightarrow$}      & \textbf{1.00} \color{equal}{$\rightarrow$}  & \textbf{0.98} \color{equal}{} & \textbf{0.98} \color{equal}{$\rightarrow$}  &      0.95 \color{equal}{$\rightarrow$}      & 0.95 \color{equal}{$\rightarrow$}           &        0.79 \color{bad}{$\searrow$}        & \textbf{0.98} \color{equal}{$\rightarrow$}  \\
	                 & 2014-12-09-13-21-02 & 2015-08-28-09-50-22 & \textbf{0.98} \color{equal}{} &      0.97 \color{equal}{$\rightarrow$}      &      0.96 \color{equal}{$\rightarrow$}      & 0.95 \color{equal}{$\rightarrow$}           &      0.91 \color{equal}{$\rightarrow$}      &      0.97 \color{equal}{$\rightarrow$}      & \textbf{0.92} \color{equal}{} &      0.87 \color{equal}{$\rightarrow$}      &      0.91 \color{equal}{$\rightarrow$}      & 0.88 \color{equal}{$\rightarrow$}           &    0.40 \color{superbad}{$\downarrow$}     &      0.89 \color{equal}{$\rightarrow$}      \\
	                 & 2014-12-09-13-21-02 & 2014-11-25-09-18-32 & \textbf{1.00} \color{equal}{} & \textbf{1.00} \color{equal}{$\rightarrow$}  &      0.99 \color{equal}{$\rightarrow$}      & 0.99 \color{equal}{$\rightarrow$}           &      0.96 \color{equal}{$\rightarrow$}      & \textbf{1.00} \color{equal}{$\rightarrow$}  & \textbf{0.99} \color{equal}{} &      0.98 \color{equal}{$\rightarrow$}      &      0.93 \color{equal}{$\rightarrow$}      & 0.98 \color{equal}{$\rightarrow$}           &        0.80 \color{bad}{$\searrow$}        &      0.98 \color{equal}{$\rightarrow$}      \\
	                 & 2014-12-09-13-21-02 & 2014-12-16-18-44-24 &     0.81 \color{equal}{}      &      0.83 \color{equal}{$\rightarrow$}      &      0.78 \color{equal}{$\rightarrow$}      & \textbf{0.85} \color{equal}{$\rightarrow$}  &        0.61 \color{bad}{$\searrow$}         &      0.76 \color{equal}{$\rightarrow$}      &     0.84 \color{equal}{}      & \textbf{0.85} \color{equal}{$\rightarrow$}  &      0.07 \color{notworking}{$\times$}      & 0.81 \color{equal}{$\rightarrow$}           &    0.60 \color{superbad}{$\downarrow$}     &      0.81 \color{equal}{$\rightarrow$}      \\ \hline\hline
	                 &                     &                     &                                                                                                                                                                                                                                         \multicolumn{12}{c}{\textbf{Multi-Matching Performance}}                                                                                                                                                                                                                                         \\ \hline
	Nordland         & fall                & spring              &     0.39 \color{equal}{}      &     0.58 \color{supergood}{$\uparrow$}      & \textbf{0.61} \color{supergood}{$\uparrow$} & 0.08 \color{notworking}{$\times$}           &     0.14 \color{superbad}{$\downarrow$}     &     0.12 \color{superbad}{$\downarrow$}     & \textbf{1.00} \color{equal}{} & \textbf{1.00} \color{equal}{$\rightarrow$}  & \textbf{1.00} \color{equal}{$\rightarrow$}  & 0.08 \color{notworking}{$\times$}           &     0.06 \color{notworking}{$\times$}      &     0.27 \color{superbad}{$\downarrow$}     \\
	                 & fall                & winter              &     0.21 \color{equal}{}      & \textbf{0.36} \color{supergood}{$\uparrow$} &        0.26 \color{good}{$\nearrow$}        & 0.05 \color{notworking}{$\times$}           &      0.01 \color{notworking}{$\times$}      &     0.10 \color{superbad}{$\downarrow$}     &     0.63 \color{equal}{}      &     0.90 \color{supergood}{$\uparrow$}      & \textbf{0.92} \color{supergood}{$\uparrow$} & 0.08 \color{notworking}{$\times$}           &    0.13 \color{superbad}{$\downarrow$}     &     0.20 \color{superbad}{$\downarrow$}     \\
	                 & spring              & winter              &     0.45 \color{equal}{}      & \textbf{0.59} \color{supergood}{$\uparrow$} &        0.50 \color{good}{$\nearrow$}        & 0.05 \color{notworking}{$\times$}           &      0.01 \color{notworking}{$\times$}      &     0.10 \color{superbad}{$\downarrow$}     &     0.52 \color{equal}{}      &     0.74 \color{supergood}{$\uparrow$}      & \textbf{0.80} \color{supergood}{$\uparrow$} & 0.07 \color{notworking}{$\times$}           &     0.76 \color{supergood}{$\uparrow$}     &     0.16 \color{superbad}{$\downarrow$}     \\
	                 & summer              & spring              &     0.43 \color{equal}{}      &     0.60 \color{supergood}{$\uparrow$}      & \textbf{0.61} \color{supergood}{$\uparrow$} & 0.09 \color{notworking}{$\times$}           &     0.17 \color{superbad}{$\downarrow$}     &     0.17 \color{superbad}{$\downarrow$}     &     0.99 \color{equal}{}      & \textbf{1.00} \color{equal}{$\rightarrow$}  & \textbf{1.00} \color{equal}{$\rightarrow$}  & 0.09 \color{notworking}{$\times$}           &        0.76 \color{bad}{$\searrow$}        &     0.28 \color{superbad}{$\downarrow$}     \\
	                 & summer              & fall                &     0.96 \color{equal}{}      & \textbf{0.97} \color{equal}{$\rightarrow$}  & \textbf{0.97} \color{equal}{$\rightarrow$}  & 0.10 \color{superbad}{$\downarrow$}         &     0.25 \color{superbad}{$\downarrow$}     &     0.24 \color{superbad}{$\downarrow$}     & \textbf{1.00} \color{equal}{} & \textbf{1.00} \color{equal}{$\rightarrow$}  & \textbf{1.00} \color{equal}{$\rightarrow$}  & 0.08 \color{notworking}{$\times$}           &    0.54 \color{superbad}{$\downarrow$}     &     0.32 \color{superbad}{$\downarrow$}     \\
	StLucia          & 100909-0845         & 190809-0845         &     0.41 \color{equal}{}      &   0.48 \color{good}{$\nearrow$}    &   0.48 \color{good}{$\nearrow$}    & 0.24 \color{superbad}{$\downarrow$}         &     0.26 \color{superbad}{$\downarrow$}     &   \textbf{0.49} \color{good}{$\nearrow$}    &     0.59 \color{equal}{}      &      0.64 \color{equal}{$\rightarrow$}      &   \textbf{0.71} \color{good}{$\nearrow$}    & 0.29 \color{superbad}{$\downarrow$}         &    0.29 \color{superbad}{$\downarrow$}     &   0.65 \color{good}{$\nearrow$}    \\
	                 & 100909-1000         & 210809-1000         &     0.47 \color{equal}{}      &   \textbf{0.54} \color{good}{$\nearrow$}    &      0.51 \color{equal}{$\rightarrow$}      & 0.21 \color{superbad}{$\downarrow$}         &     0.27 \color{superbad}{$\downarrow$}     &   0.53 \color{good}{$\nearrow$}    &     0.57 \color{equal}{}      &        0.67 \color{good}{$\nearrow$}        &   \textbf{0.71} \color{good}{$\nearrow$}    & 0.32 \color{superbad}{$\downarrow$}         &    0.29 \color{superbad}{$\downarrow$}     &   0.65 \color{good}{$\nearrow$}    \\
	                 & 100909-1210         & 210809-1210         &     0.51 \color{equal}{}      &   \textbf{0.61} \color{good}{$\nearrow$}    &        0.60 \color{good}{$\nearrow$}        & 0.31 \color{superbad}{$\downarrow$}         &      0.07 \color{notworking}{$\times$}      & 0.53 \color{equal}{$\rightarrow$}  &     0.54 \color{equal}{}      &        0.61 \color{good}{$\nearrow$}        &   \textbf{0.65} \color{good}{$\nearrow$}    & 0.31 \color{superbad}{$\downarrow$}         &    0.20 \color{superbad}{$\downarrow$}     &   0.64 \color{good}{$\nearrow$}    \\
	                 & 100909-1410         & 190809-1410         &     0.38 \color{equal}{}      &     0.56 \color{supergood}{$\uparrow$}      & \textbf{0.57} \color{supergood}{$\uparrow$} & 0.15 \color{superbad}{$\downarrow$}         &     0.27 \color{superbad}{$\downarrow$}     &   0.46 \color{good}{$\nearrow$}    &     0.61 \color{equal}{}      &      0.61 \color{equal}{$\rightarrow$}      & 0.64 \color{equal}{$\rightarrow$}  & 0.34 \color{superbad}{$\downarrow$}         &    0.30 \color{superbad}{$\downarrow$}     & \textbf{0.66} \color{equal}{$\rightarrow$}  \\
	                 & 110909-1545         & 180809-1545         &     0.27 \color{equal}{}      &      0.29 \color{equal}{$\rightarrow$}      &   0.32 \color{good}{$\nearrow$}    & 0.15 \color{superbad}{$\downarrow$}         &      0.05 \color{notworking}{$\times$}      & \textbf{0.41} \color{supergood}{$\uparrow$} &     0.60 \color{equal}{}      &   \textbf{0.68} \color{good}{$\nearrow$}    &      0.65 \color{equal}{$\rightarrow$}      & 0.36 \color{superbad}{$\downarrow$}         &    0.31 \color{superbad}{$\downarrow$}     &   0.67 \color{good}{$\nearrow$}    \\
	CMU              & 20110421            & 20100901            &     0.73 \color{equal}{}      & \textbf{0.76} \color{equal}{$\rightarrow$}  &        0.65 \color{bad}{$\searrow$}         & 0.49 \color{superbad}{$\downarrow$}         &     0.10 \color{superbad}{$\downarrow$}     &     0.40 \color{superbad}{$\downarrow$}     &     0.44 \color{equal}{}      &   \textbf{0.52} \color{good}{$\nearrow$}    &      0.43 \color{equal}{$\rightarrow$}      & 0.32 \color{superbad}{$\downarrow$}         &     0.09 \color{notworking}{$\times$}      &     0.29 \color{superbad}{$\downarrow$}     \\
	                 & 20110421            & 20100915            &     0.77 \color{equal}{}      &     0.56 \color{superbad}{$\downarrow$}     & \textbf{0.80} \color{equal}{$\rightarrow$}  & 0.32 \color{superbad}{$\downarrow$}         &      0.04 \color{notworking}{$\times$}      &     0.39 \color{superbad}{$\downarrow$}     &     0.59 \color{equal}{}      & \textbf{0.64} \color{equal}{$\rightarrow$}  &        0.48 \color{bad}{$\searrow$}         & 0.31 \color{superbad}{$\downarrow$}         &    0.14 \color{superbad}{$\downarrow$}     &     0.32 \color{superbad}{$\downarrow$}     \\
	                 & 20110421            & 20101221            &     0.56 \color{equal}{}      & \textbf{0.85} \color{supergood}{$\uparrow$} &      0.61 \color{equal}{$\rightarrow$}      & 0.45 \color{bad}{$\searrow$}                &     0.29 \color{superbad}{$\downarrow$}     &     0.31 \color{superbad}{$\downarrow$}     &     0.34 \color{equal}{}      & \textbf{0.69} \color{supergood}{$\uparrow$} &     0.68 \color{supergood}{$\uparrow$}      & 0.33 \color{equal}{$\rightarrow$}           &     0.31 \color{equal}{$\rightarrow$}      &        0.28 \color{bad}{$\searrow$}         \\
	                 & 20110421            & 20110202            &     0.61 \color{equal}{}      &   \textbf{0.68} \color{good}{$\nearrow$}    &   \textbf{0.68} \color{good}{$\nearrow$}    & 0.52 \color{bad}{$\searrow$}                &      0.03 \color{notworking}{$\times$}      &        0.51 \color{bad}{$\searrow$}         &     0.33 \color{equal}{}      &        0.40 \color{good}{$\nearrow$}        &   \textbf{0.41} \color{good}{$\nearrow$}    & 0.24 \color{superbad}{$\downarrow$}         &     0.01 \color{notworking}{$\times$}      &   \textbf{0.41} \color{good}{$\nearrow$}    \\
	GardensPoint     & day-right           & day-left            &     0.97 \color{equal}{}      &      0.97 \color{equal}{$\rightarrow$}      &      0.97 \color{equal}{$\rightarrow$}      & \textbf{1.00} \color{equal}{$\rightarrow$}  &      0.98 \color{equal}{$\rightarrow$}      & 0.97 \color{equal}{$\rightarrow$}  &     0.58 \color{equal}{}      &        0.71 \color{good}{$\nearrow$}        &        0.68 \color{good}{$\nearrow$}        & \textbf{0.87} \color{supergood}{$\uparrow$} &     0.76 \color{supergood}{$\uparrow$}     & 0.59 \color{equal}{$\rightarrow$}  \\
	                 & day-right           & night-right         &     0.51 \color{equal}{}      &        0.63 \color{good}{$\nearrow$}        &     0.66 \color{supergood}{$\uparrow$}      & 0.70 \color{supergood}{$\uparrow$}          & \textbf{0.77} \color{supergood}{$\uparrow$} & 0.56 \color{equal}{$\rightarrow$}  &     0.51 \color{equal}{}      &        0.62 \color{good}{$\nearrow$}        &        0.58 \color{good}{$\nearrow$}        & \textbf{0.86} \color{supergood}{$\uparrow$} &     0.64 \color{supergood}{$\uparrow$}     & 0.65 \color{supergood}{$\uparrow$} \\
	                 & day-left            & night-right         &     0.40 \color{equal}{}      &     0.56 \color{supergood}{$\uparrow$}      &     0.54 \color{supergood}{$\uparrow$}      & \textbf{0.78} \color{supergood}{$\uparrow$} &     0.60 \color{supergood}{$\uparrow$}      & 0.42 \color{equal}{$\rightarrow$}  &     0.11 \color{equal}{}      &     0.30 \color{supergood}{$\uparrow$}      & \textbf{0.37} \color{supergood}{$\uparrow$} & 0.35 \color{supergood}{$\uparrow$}          &     0.30 \color{supergood}{$\uparrow$}     & 0.16 \color{supergood}{$\uparrow$} \\
	Oxford           & 2014-12-09-13-21-02 & 2015-05-19-14-06-38 & \textbf{0.78} \color{equal}{} & \textbf{0.78} \color{equal}{$\rightarrow$}  & \textbf{0.78} \color{equal}{$\rightarrow$}  & 0.38 \color{superbad}{$\downarrow$}         &     0.24 \color{superbad}{$\downarrow$}     &     0.42 \color{superbad}{$\downarrow$}     &     0.24 \color{equal}{}      & \textbf{0.41} \color{supergood}{$\uparrow$} &     0.39 \color{supergood}{$\uparrow$}      & 0.19 \color{bad}{$\searrow$}                &     0.02 \color{notworking}{$\times$}      &   0.29 \color{good}{$\nearrow$}    \\
	                 & 2014-12-09-13-21-02 & 2015-08-28-09-50-22 & \textbf{0.60} \color{equal}{} &        0.45 \color{bad}{$\searrow$}         &     0.32 \color{superbad}{$\downarrow$}     & 0.16 \color{superbad}{$\downarrow$}         &     0.27 \color{superbad}{$\downarrow$}     &     0.29 \color{superbad}{$\downarrow$}     &     0.11 \color{equal}{}      &     0.16 \color{supergood}{$\uparrow$}      & \textbf{0.23} \color{supergood}{$\uparrow$} & 0.12 \color{equal}{$\rightarrow$}           &     0.00 \color{notworking}{$\times$}      & 0.17 \color{supergood}{$\uparrow$} \\
	                 & 2014-12-09-13-21-02 & 2014-11-25-09-18-32 & \textbf{0.87} \color{equal}{} & \textbf{0.87} \color{equal}{$\rightarrow$}  &        0.76 \color{bad}{$\searrow$}         & 0.57 \color{superbad}{$\downarrow$}         &      0.06 \color{notworking}{$\times$}      &     0.60 \color{superbad}{$\downarrow$}     &     0.42 \color{equal}{}      &   \textbf{0.49} \color{good}{$\nearrow$}    &        0.47 \color{good}{$\nearrow$}        & 0.24 \color{superbad}{$\downarrow$}         &     0.01 \color{notworking}{$\times$}      &        0.35 \color{bad}{$\searrow$}         \\
	                 & 2014-12-09-13-21-02 & 2014-12-16-18-44-24 &     0.55 \color{equal}{}      &   \textbf{0.62} \color{good}{$\nearrow$}    &      0.60 \color{equal}{$\rightarrow$}      & 0.10 \color{superbad}{$\downarrow$}         &      0.02 \color{notworking}{$\times$}      &     0.12 \color{superbad}{$\downarrow$}     &     0.07 \color{equal}{}      & \textbf{0.27} \color{supergood}{$\uparrow$} &      0.01 \color{notworking}{$\times$}      & 0.07 \color{notworking}{$\times$}           &     0.02 \color{notworking}{$\times$}      & 0.12 \color{supergood}{$\uparrow$}
\end{tabular}
}
\vspace{-0.1cm}
\end{table*}
\unboldmath

The multi-matching performance is often improved by our proposed methods, because of their ability to derive multiple matches from the intra-database similarities $S^{DB}$ while ignoring similar looking image pairs of different places.
OPR, HNSW and EPR-PR without using $S^{DB}$ fail on many datasets.
HNSW can achieve high performance on some datasets with few loops and no stops, while FLANN~\cite{flann} as an ANN-based method achieved only low performance in the literature on  place recognition~\cite{Vysotska2016}.

EPR-PR without an exploitation of intra-database similarities (EPR-PR w/o $S^{DB}$) fails on the multi-matching task while it shows a surprisingly well single-matching performance.
In conclusion, our proposed methods achieved the most performance gains, and performed best on most datasets for both the single- and multi-matching use-case.

\subsection{Percentage of evaluated image pairs}
\begin{figure}[t]
    \centering
    \includegraphics[width=0.49\linewidth]{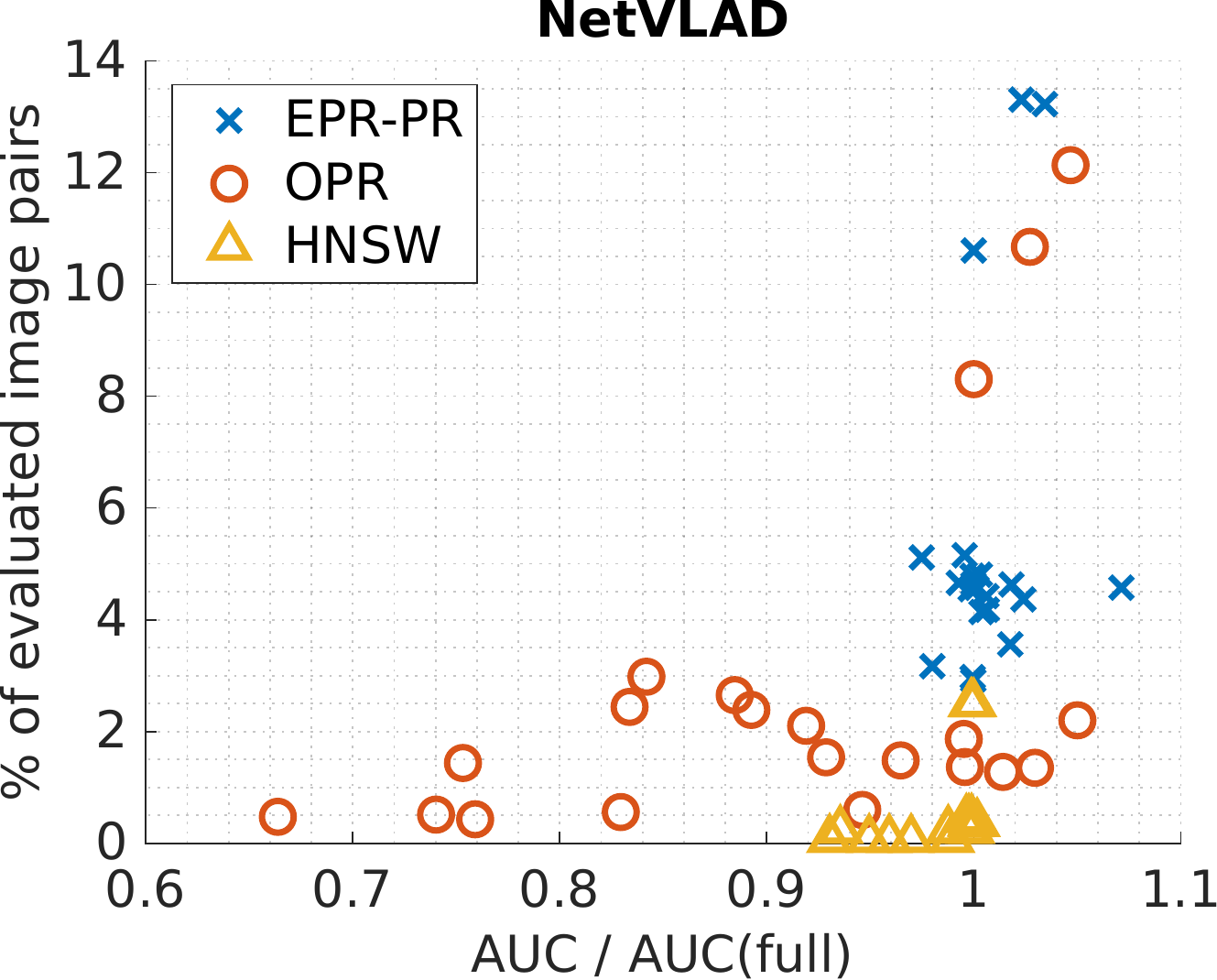}
    \includegraphics[width=0.49\linewidth]{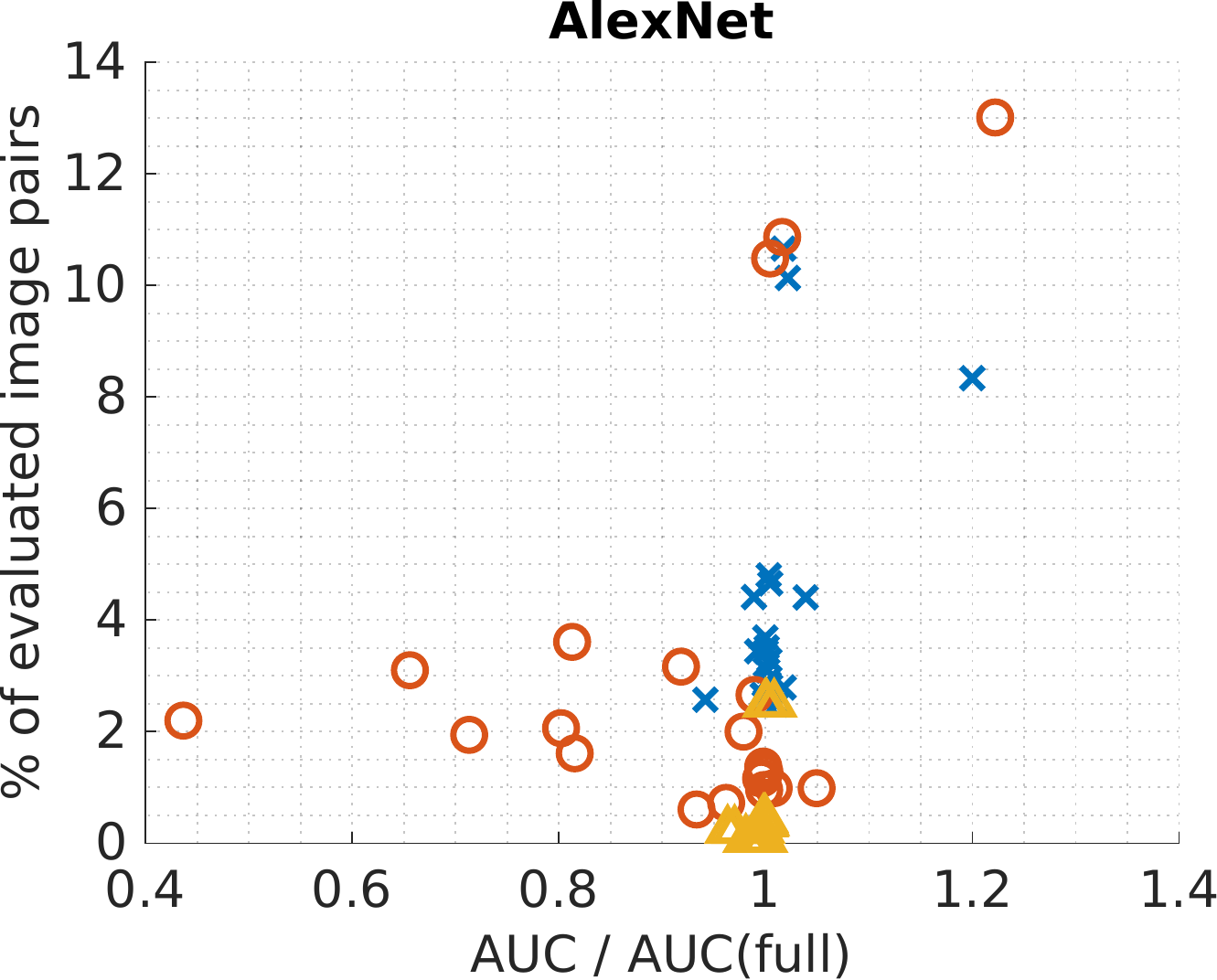}
    \includegraphics[width=0.49\linewidth]{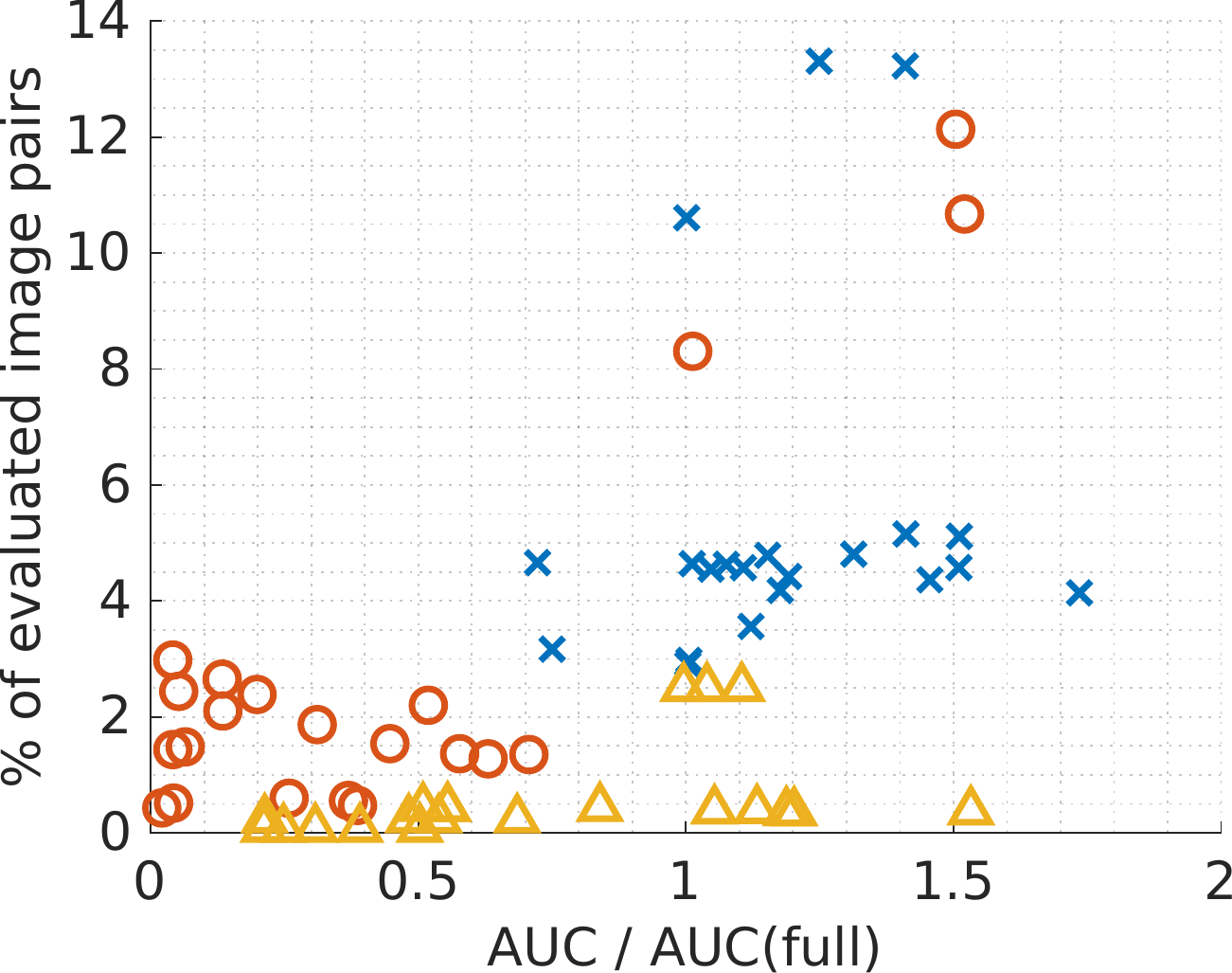}
    \includegraphics[width=0.49\linewidth]{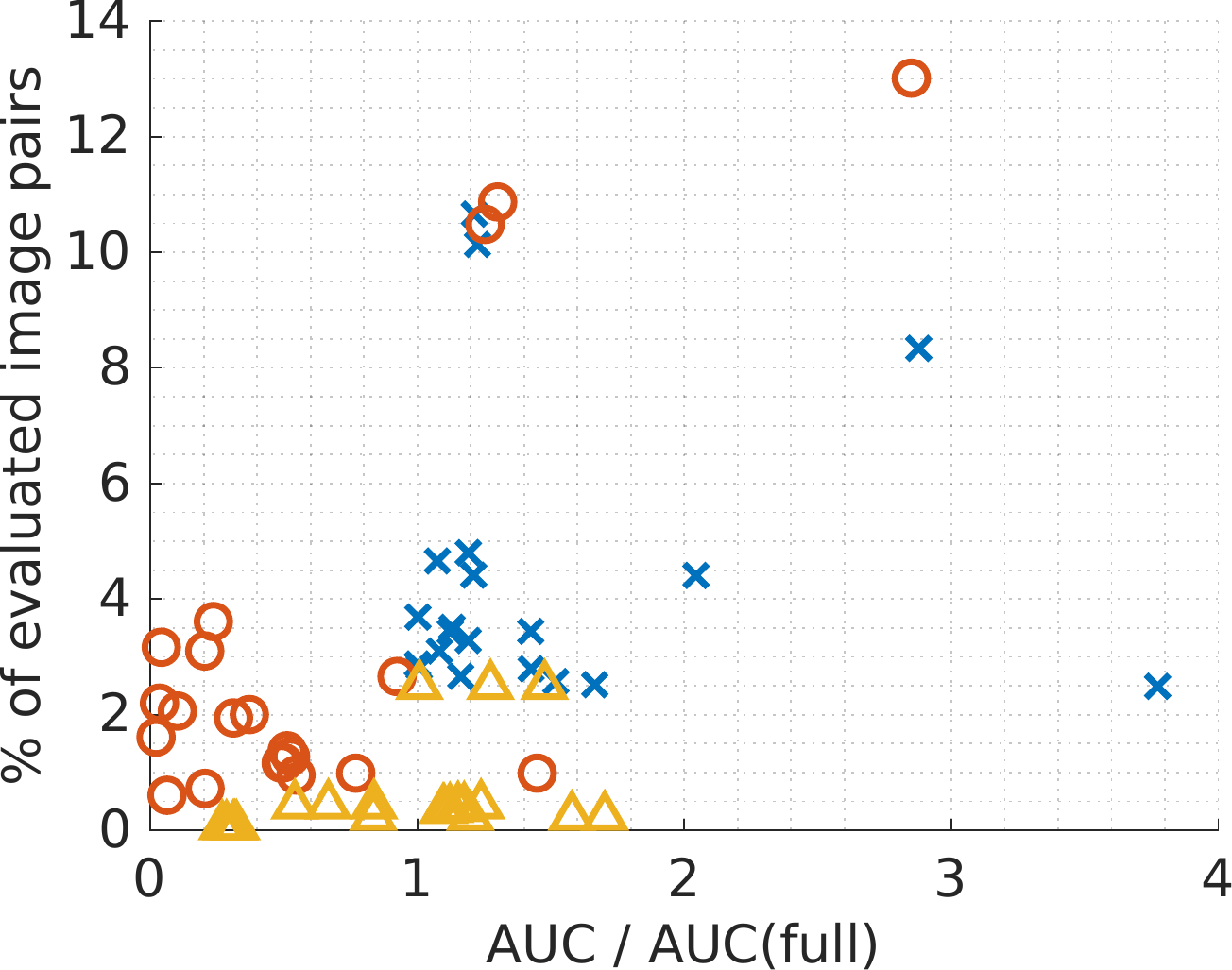}
    
    \vspace{-0.25cm}
    
    \caption{Relative performance improvement compared to full image comparisons vs percentage of evaluated image pairs over all datasets from Table~\ref{tab:AUC} and \ref{tab:density} for single-matching performance (top) and multi-matching performance (bottom)}
    \label{fig:density-vs-AUC}
    
    \vspace{-0.2cm}
\end{figure}

Table~\ref{tab:density} shows the corresponding percentages of evaluated image pairs over all datasets and methods.
Fig.~\ref{fig:density-vs-AUC} illustrates the relation between the achieved performance gain over full image comparisons and the percentage of evaluated image pairs for our best performing method EPR-PR as well as the two methods OPR and HNSW.

Our method EPR-PR achieves a good tradeoff between performance gain or preservation and percentage of comparisons for single and multi matching, while OPR partially fails.
HNSW by design achieves the lowest percentage, but partially fails on multi-matching.

\boldmath
\begin{table*}[t]
\centering
\caption{Percentage of evaluated image pairs between database and query. The ground truth (GT) shows the percentage of required (min) and allowed (max) matches (allowed matches due to small visual overlaps between consecutive images)}\vspace{-0.3cm}

\label{tab:density}
\resizebox{1\textwidth}{!}{%
\begin{tabular}{lll||cc||c|cc|P{1.25cm}|cc||c|cc|P{1.25cm}|cc}
	\hline
	                 &                     &                     & \multicolumn{2}{c||}{\textbf{GT}} &                                     \multicolumn{6}{c||}{\textbf{NetVLAD}}                                      &                                      \multicolumn{6}{c}{\textbf{AlexNet}}                                       \\ 
                     \textbf{Dataset} & \textbf{Database} & \textbf{Query} &\textbf{min} &\textbf{max} &\textbf{full} &\textbf{EPR-PR}  &\textbf{EPR-ER}&\textbf{EPR-PR w/o $S^{DB}$}&\textbf{OPR}&\textbf{HNSW} &\textbf{full} &\textbf{EPR-PR}  &\textbf{EPR-ER} &\textbf{EPR-PR w/o $S^{DB}$}&\textbf{OPR} &\textbf{HNSW}\\ \hline\hline
	Nordland          & fall                & spring              &     1.27     &        2.11        &      100      &      5.11       &      4.36       & 1.20                         &     0.56     &     0.07      &      100      &      2.87       &      4.23       & 1.18                         &     0.61     &     0.07      \\
	                  & fall                & winter              &     1.27     &        2.11        &      100      &      4.14       &      3.94       & 1.23                         &     0.51     &     0.07      &      100      &      2.79       &      5.42       & 1.18                         &     0.72     &     0.07      \\
	                  & spring              & winter              &     1.27     &        2.11        &      100      &      4.81       &      5.43       & 1.22                         &     0.43     &     0.07      &      100      &      3.44       &      11.80      & 1.18                         &     0.99     &     0.07      \\
	                  & summer              & spring              &     1.27     &        2.11        &      100      &      5.15       &      4.33       & 1.22                         &     0.47     &     0.07      &      100      &      2.82       &      2.83       & 1.19                         &     0.99     &     0.07      \\
	                  & summer              & fall                &     1.27     &        2.11        &      100      &      4.64       &      3.72       & 1.21                         &     0.60     &     0.07      &      100      &      2.88       &      2.04       & 1.17                         &     0.95     &     0.07      \\
	StLucia           & 100909-0845         & 190809-0845         &     0.17     &        1.24        &      100      &      4.18       &      3.26       & 2.06                         &     1.28     &     0.35      &      100      &      3.10       &      2.27       & 2.01                         &     1.16     &     0.35      \\
	                  & 100909-1000         & 210809-1000         &     0.16     &        1.26        &      100      &      4.79       &      3.90       & 2.18                         &     1.36     &     0.39      &      100      &      3.28       &      2.51       & 2.11                         &     1.37     &     0.39      \\
	                  & 100909-1210         & 210809-1210         &     0.14     &        1.29        &      100      &      4.42       &      3.57       & 2.21                         &     2.10     &     0.39      &      100      &      3.43       &      2.67       & 2.20                         &     2.00     &     0.39      \\
	                  & 100909-1410         & 190809-1410         &     0.17     &        1.28        &      100      &      4.36       &      3.38       & 2.12                         &     1.35     &     0.36      &      100      &      3.69       &      2.87       & 2.01                         &     1.17     &     0.36      \\
	                  & 110909-1545         & 180809-1545         &     0.14     &        1.24        &      100      &      4.63       &      3.69       & 2.17                         &     2.39     &     0.37      &      100      &      3.51       &      3.38       & 2.03                         &     1.28     &     0.37      \\
	CMU               & 20110421            & 20100901            &     0.59     &        1.97        &      100      &      4.55       &      4.06       & 2.11                         &     2.65     &     0.43      &      100      &      4.80       &      5.41       & 2.21                         &     3.10     &     0.43      \\
	                  & 20110421            & 20100915            &     0.47     &        1.66        &      100      &      4.66       &      4.66       & 2.04                         &     2.44     &     0.43      &      100      &      4.66       &      5.23       & 2.11                         &     3.61     &     0.43      \\
	                  & 20110421            & 20101221            &     0.57     &        1.96        &      100      &      4.57       &      4.19       & 2.20                         &     2.20     &     0.43      &      100      &      4.41       &      4.79       & 2.20                         &     2.66     &     0.43      \\
	                  & 20110421            & 20110202            &     0.32     &        1.97        &      100      &      4.57       &      3.89       & 2.15                         &     2.98     &     0.43      &      100      &      4.41       &      6.34       & 2.27                         &     3.17     &     0.43      \\
	GardensPoint      & day-right           & day-left            &     0.50     &        8.32        &      100      &      10.61      &      9.66       & 6.60                         &     8.31     &     2.50      &      100      &      10.66      &      13.20      & 7.35                         &    10.87     &     2.50      \\
	                  & day-right           & night-right         &     0.50     &        8.32        &      100      &      13.31      &      14.06      & 7.78                         &    10.67     &     2.50      &      100      &      10.14      &      16.53      & 7.30                         &    10.48     &     2.50      \\
	                  & day-left            & night-right         &     0.50     &        8.32        &      100      &      13.23      &      15.40      & 7.69                         &    12.14     &     2.50      &      100      &      8.34       &      31.08      & 7.02                         &    13.01     &     2.50      \\
	Oxford            & 2014-12-09-13-21-02 & 2015-05-19-14-06-38 &     0.13     &        0.93        &      100      &      2.91       &      2.02       & 1.60                         &     1.87     &     0.23      &      100      &      2.52       &      1.56       & 1.60                         &     2.06     &     0.23      \\
	                  & 2014-12-09-13-21-02 & 2015-08-28-09-50-22 &     0.07     &        0.82        &      100      &      3.17       &      2.59       & 1.63                         &     1.54     &     0.23      &      100      &      2.57       &      2.93       & 1.59                         &     2.20     &     0.23      \\
	                  & 2014-12-09-13-21-02 & 2014-11-25-09-18-32 &     0.07     &        0.93        &      100      &      2.97       &      2.54       & 1.58                         &     1.48     &     0.23      &      100      &      2.66       &      1.38       & 1.61                         &     1.61     &     0.23      \\
	                  & 2014-12-09-13-21-02 & 2014-12-16-18-44-24 &     0.14     &        0.92        &      100      &      3.56       &      2.77       & 1.69                         &     1.44     &     0.23      &      100      &      2.49       &      1.40       & 1.64                         &     1.94     &     0.23
\end{tabular}
}
\vspace{-0.3cm}
\end{table*}
\unboldmath

\subsection{Influence of relocalization}\label{sec:reloc_problem}
\begin{figure}[t]
    \centering
    \includegraphics[width=0.325\linewidth]{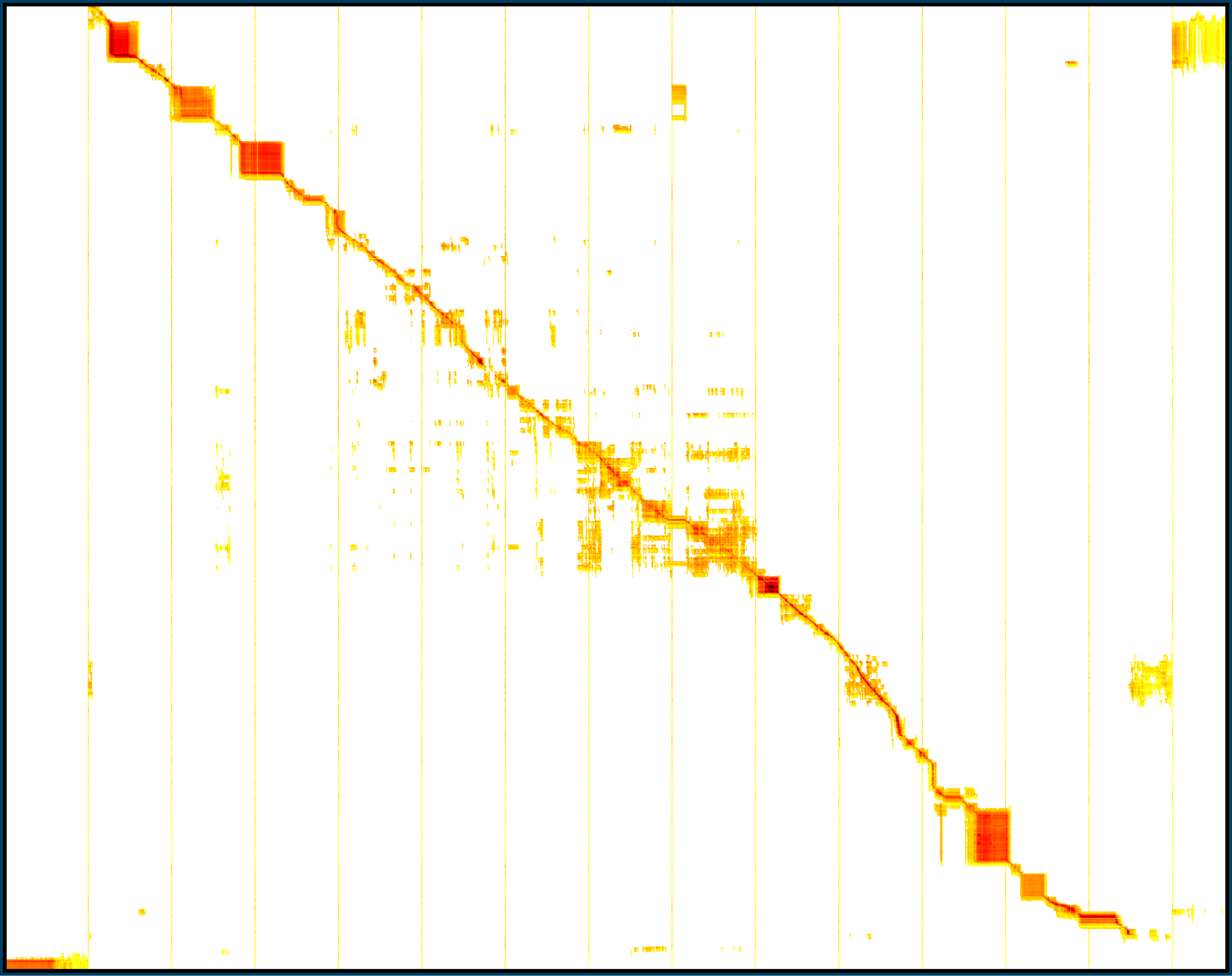}
    \includegraphics[width=0.325\linewidth]{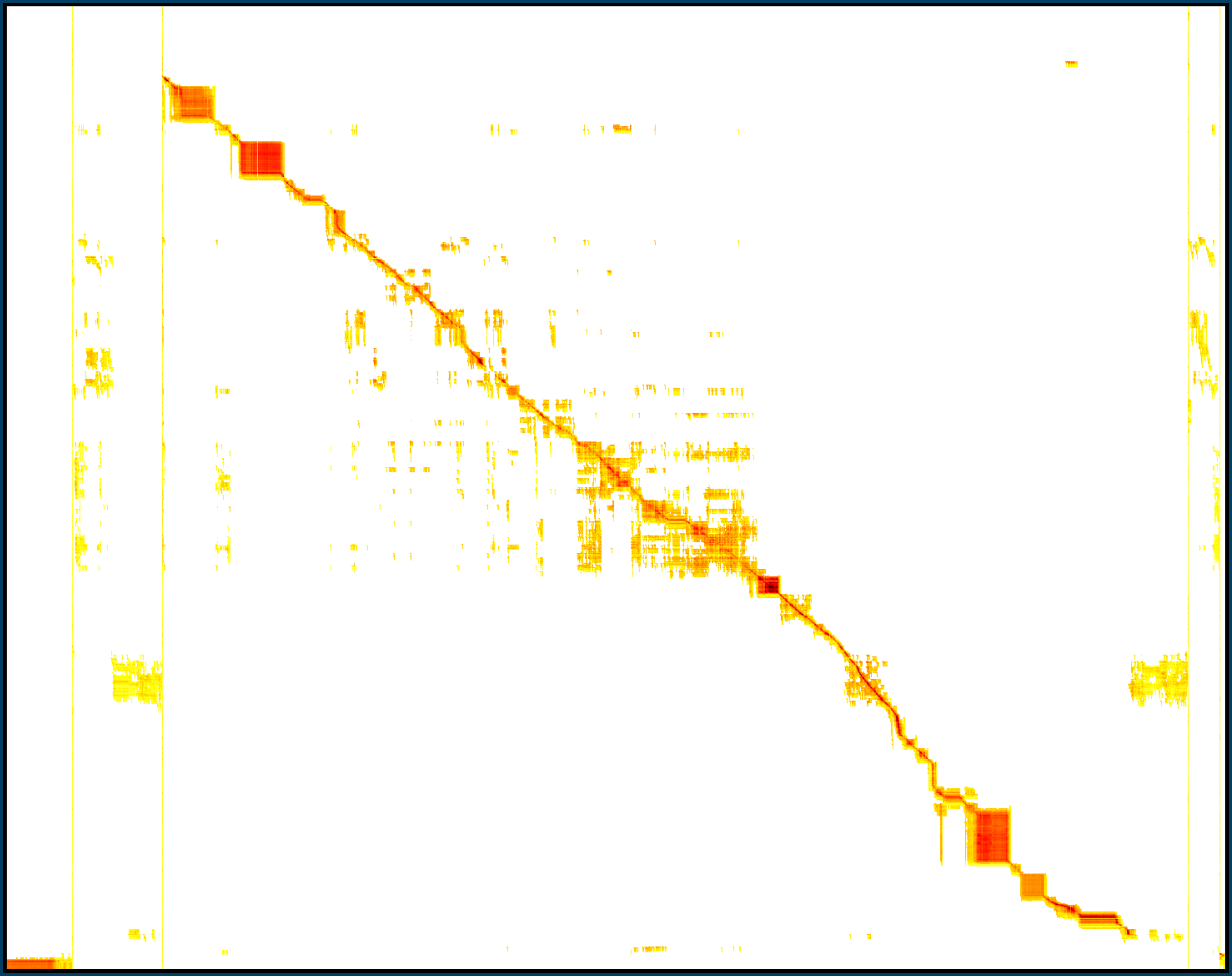}
    \includegraphics[width=0.325\linewidth]{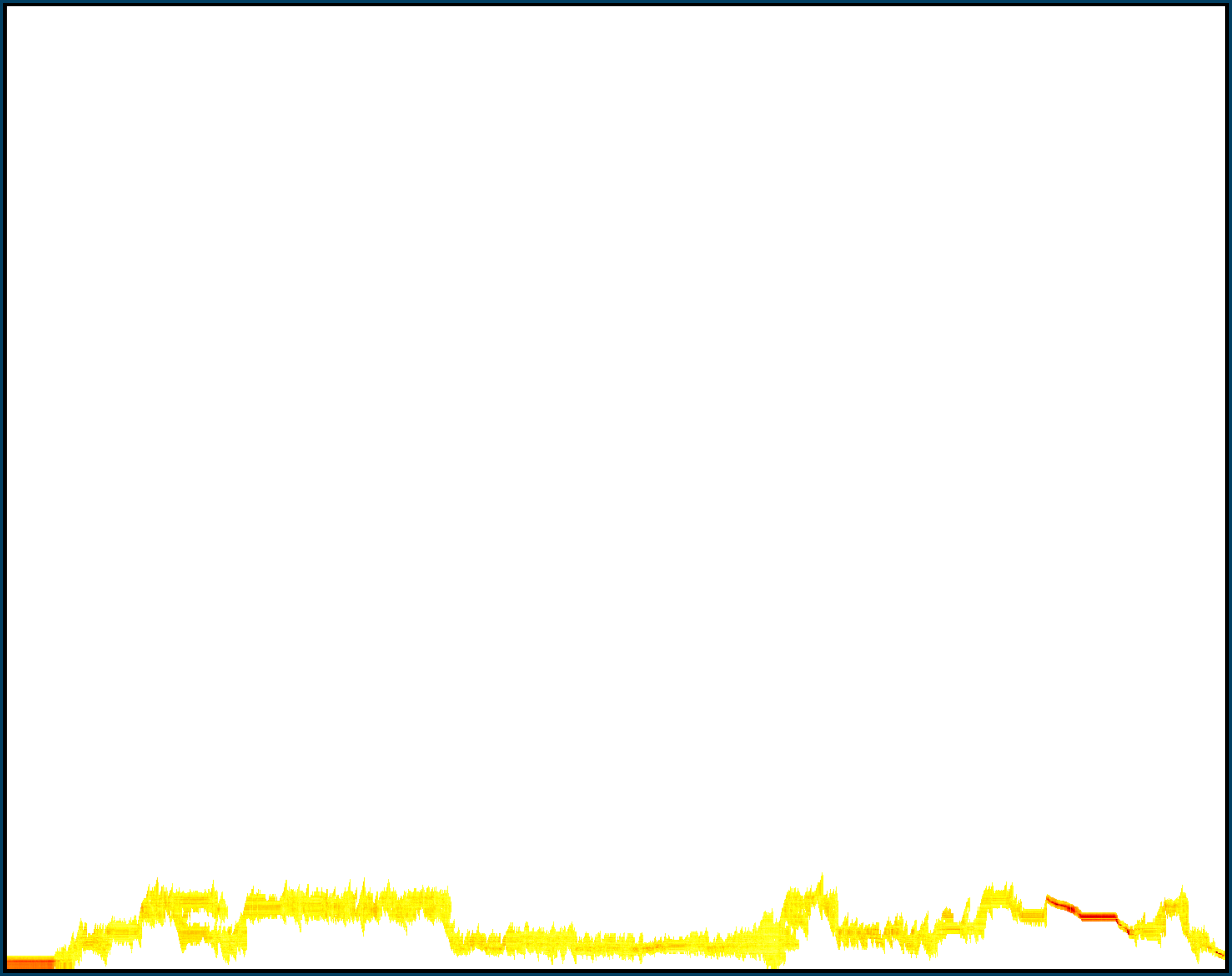}
    
    \vspace{-0.2cm}
    
    \caption{Visualization of the exploratory query problem in $S\in\mathbb{R}^{|DB|\times|Q|}$: Without a relocalization strategy, sequence-based methods are likely to fail if the first or intermediate query images do not show places that are represented in the database. While our methods EPR-PR (left) and EPR-ER (mid) can recover due to relocalization, OPR (right) cannot}
    \label{fig:exploration_problem}
    
    \vspace{-0.3cm}
\end{figure}

Fig.~\ref{fig:exploration_problem} shows the problem of exploration during the query run and the benefit of relocalization: The shown dataset starts with exploratory queries.
While EPR-PR can recover right after the beginning of the actual sequence, EPR-ER cannot immediately detect the need for a relocalization, but after a short period of time.
OPR without a relocalization procedure completely fails to recover from the sequence loss.

\subsection{Runtime}
We implemented our proposed methods in Matlab and measured the runtimes on a notebook with Intel i7-8550U CPU.
On the largest datasets with $6862$ database and $6862$ query images, our method EPR-PR required $7$sec without the runtime for descriptor computations and the in average $330$ image comparisons per query.
On the same dataset, OPR required $142$sec without descriptor computations and image comparisons despite its C++ implementation.

\section{CONCLUSION}\label{sec:discussion}

The experimental evaluation in this paper showed that the two state-of-the-art methods OPR and HNSW can considerably reduce the number of comparisons. 
However, each approach also revealed cases where the place recognition performance decreased compared to an exhaustive full comparison. 
This became obvious in multi-matching scenarios which are important in online scenarios, e.g., for SLAM.
In this paper, we proposed a new method, EPR, for efficient place recognition.
The approach exploits structural knowledge, which is inherent to many practical place recognition setups, for a reduction of descriptor comparisons.
Our experiments demonstrated that a single set of parameters could be successfully used for a wide range of datasets. 
This is partially due to the ability to auto-tune required thresholds.
The comparison with the state-of-the-art approaches showed a good balance between performance and efficiency.
Although, there still exists rare cases where the performance decreased, in significantly more cases, the presented approach was able to even improve the place recognition performance compared to a full comparison. 
This is based on the ability to use structural information to prevent the evaluation of potential false positive matchings.
A key element of the presented approach to use structural knowledge is the ability to relocalize and to recognize when this becomes necessary. We presented two different relocalization strategies.
In principle, they also allow the combination with ANN-approaches for application on very large scale datasets.
A particularly interesting open research question are more robust event-based relocalization strategies for immediate relocalization.

\end{document}